\documentclass[11pt,a4paper]{article}
\usepackage[hyperref]{emnlp2020}
\usepackage{times}
\usepackage{latexsym}
\usepackage{program_synthesis}
\usepackage{aplcomments}

\newcommand{\Real}{\mathbb{R}}

\newcommand{\rom}[1]{%
	\textup{\uppercase\expandafter{\romannumeral#1}}%
}

\newcommand{\eat}[1]{\ignorespaces}

\newcolumntype{L}[1]{>{\raggedright\let\newline\\\arraybackslash\hspace{0pt}}m{#1}}
\newcolumntype{C}[1]{>{\centering\let\newline\\\arraybackslash\hspace{0pt}}m{#1}}
\newcolumntype{R}[1]{>{\raggedleft\let\newline\\\arraybackslash\hspace{0pt}}m{#1}}

\def\|#1|{\mathid{#1}}
\newcommand{\mathid}[1]{\ensuremath{\mathit{#1}}}
\def\<#1>{\codeid{#1}}
\protected\def\codeid#1{\ifmmode{\mbox{\smaller\ttfamily{#1}}}\else{\smaller\ttfamily
		#1}\fi}

\newcommand{\question}{Q}

\newcommand{\sColumnSet}{\mathcal{C}}
\newcommand{\sColumn}{c}
\newcommand{\sTableSet}{\mathcal{T}}
\newcommand{\sTable}{t}
\newcommand{\sVocab}{\mathcal{V}}
\newcommand{\sPicklist}{\mathcal{P}}

\newcommand{\sql}{Y}
\newcommand{\type}{\tau}
\newcommand{\schema}{\mathcal{S}}

\definecolor{light-gray}{rgb}{.902, .902, .902}
\newcommenter{chenglong}{1.0,0.4,1.0} 
\newcommenter{mde}{0.4,0.4,1.0} 
\newcommenter{vic}{1.0,0.4,0.4} 
\newcommenter{luke}{0.4,1,0.4} 
\newcommenter{deric}{1.0,1.0,0.4} 

\newcommand{\pgen}{p_{\text{gen}}}
\newcommand{\pout}{p_{\text{out}}}

\newcommand{\hide}[1]{}

\newtheorem{lemma}{Lemma}

\aclfinalcopy 
\def\aclpaperid{3590} 

\newcommand{\reducedstrut}{\vrule width 0pt height .9\ht\strutbox depth .9\dp\strutbox\relax}
\newcommand{\colbox}[2]{  \begingroup
  \setlength{\fboxsep}{0pt}%
  \colorbox{#1}{\reducedstrut#2\/}%
  \endgroup}
\definecolor{cspiderback}{rgb}{0.3, 0.47, 0.67}
\definecolor{cspider}{rgb}{1, 1, 1}
\definecolor{cwikisqlback}{rgb}{0.62, 0.79, 0.91}
\definecolor{cwikisql}{rgb}{0, 0, 0}

\newcommand{\tabert}{TaBERT\xspace}

\newcommand{\Ours}{BRIDGE\xspace}

\title{Bridging Textual and Tabular Data for \\ Cross-Domain Text-to-SQL Semantic Parsing}

\author{Xi Victoria Lin \qquad
  Richard Socher \qquad
  Caiming Xiong \\
  Salesforce Research \\
  {\tt \{xilin,rsocher,cxiong\}@salesforce.com}
}
\date{}

\begin{document}
\maketitle

\fbox{%
  \parbox{0.4\textwidth}{
  \small
    This work first appeared in Findings of EMNLP 2020. 
    Here we report the performance and analysis of the BRIDGE model using BERT-large, on both the Spider and WikiSQL datasets. We also extend the discussion on model variance and present an ensemble model that significantly outperforms single models on Spider.
  }%
}
\vspace{2em}

\begin{abstract}
We present \Ours, 
a powerful sequential architecture for modeling dependencies between natural language questions and relational databases in cross-DB semantic parsing. 
\Ours represents the question and DB schema in a tagged sequence where a subset of the fields are augmented with cell values mentioned in the question. 
The hybrid sequence is encoded by BERT with minimal subsequent layers and the text-DB contextualization is realized via the fine-tuned deep attention in BERT.
Combined with a pointer-generator decoder with schema-consistency driven search space pruning, \Ours attained state-of-the-art performance on popular cross-DB text-to-SQL benchmarks, Spider 
(71.1\% dev, 67.5\% test with ensemble model) and WikiSQL (92.6\% dev, 91.9\% test). 
Our analysis shows that \Ours effectively captures the desired cross-modal dependencies and has the potential to generalize to more text-DB related tasks.
Our implementation is available at \url{https://github.com/salesforce/TabularSemanticParsing}.

\end{abstract}
\section{Introduction}
Text-to-SQL semantic parsing addresses the problem of mapping natural language utterances to executable relational DB queries.
Early work in this area focus on training and testing the semantic parser on a single  DB~\cite{DBLP:conf/naacl/HemphillGD90,DBLP:conf/naacl/DahlBBFHPPRS94,DBLP:conf/aaai/ZelleM96,DBLP:conf/uai/ZettlemoyerC05,DBLP:conf/acl/DongL16}. 
However, DBs are widely used in many domains and developing a semantic parser for each individual DB is unlikely to scale in practice.

\begin{figure}[t]
	\centering
	\includegraphics[width=.48\textwidth]{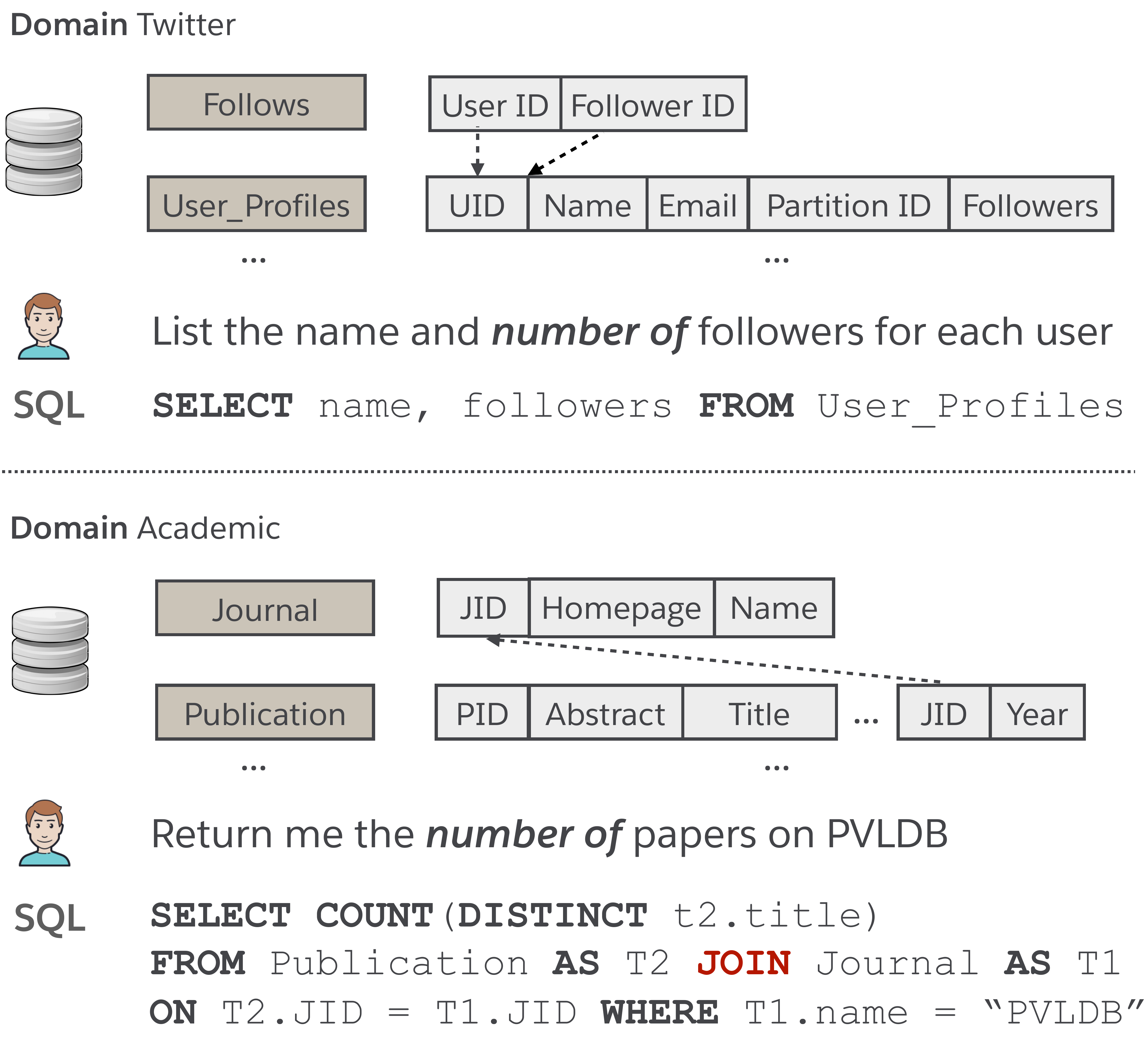}
	\caption{Two questions from the Spider dataset with similar intent resulted in completely different SQL logical forms on two DBs. In cross-DB text-to-SQL semantic parsing, the interpretation of a natural language question is strictly grounded in the underlying relational DB schema.}
	\label{fig:spider_example}
\end{figure}

More recently, large-scale datasets consisting of hundreds of DBs and the corresponding question-SQL pairs 
have been released~\cite{DBLP:conf/emnlp/YuYYZWLR18,DBLP:journals/corr/abs-1709-00103,DBLP:conf/acl/YuZYTLLELPCJDPS19,DBLP:journals/corr/abs-1909-05378} to encourage the development of semantic parsers that can work well across different DBs~\cite{DBLP:conf/acl/GuoZGXLLZ19,DBLP:conf/emnlp/BoginGB19,DBLP:conf/emnlp/YuYYZWLR19,Wang2019RATSQLRS,Suhr2020,DBLP:journals/corr/abs-2004-03125}.
The setup is challenging
as it requires the model to 
interpret a question conditioned on a 
relational DB unseen during training and accurately express the question intent via SQL logic.
Consider the two examples shown in Figure~\ref{fig:spider_example}, both questions have the intent to count, but the corresponding SQL queries are drastically different due to differences in the target DB schema. 
As a result, cross-DB text-to-SQL semantic parsers cannot trivially memorize seen SQL patterns, but instead has to accurately model the natural language question, the target DB structure, and the contextualization of both.

State-of-the-art cross-DB text-to-SQL semantic parsers 
adopt the following design principles to address the aforementioned challenges. First, the question and schema representation 
are contextualized with each other~\cite{DBLP:journals/corr/abs-1902-01069,DBLP:conf/acl/GuoZGXLLZ19,Wang2019RATSQLRS,DBLP:journals/corr/abs-2005-08314}. 
Second, 
pre-trained language models (LMs) such as BERT~\cite{DBLP:conf/naacl/DevlinCLT19} and RoBERTa~\cite{DBLP:journals/corr/abs-1907-11692} can significantly boost parsing accuracy by 
enhancing generalization over natural language variations and capturing long-term dependencies~\cite{DBLP:journals/corr/abs-2010-12725}. 
Third, 
as much as data privacy allows, leveraging available DB content improves understanding of the DB schema~\cite{DBLP:conf/emnlp/BoginGB19,Wang2019RATSQLRS,DBLP:journals/corr/abs-2005-08314}. Consider the second example in Figure~\ref{fig:spider_example}, knowing ``PLVDB'' is a value of the field \<Journal.Name> helps the model to generate the \<WHERE> condition. 

We introduce \Ours, a powerful sequential 
text-DB encoding framework assembling the three design principles mentioned above.
\Ours represents the relational DB schema as a tagged sequence concatenated to the question. 
In contrast to previous work which proposed 
task-specific layers for modeling the DB schema~\cite{DBLP:conf/acl/BoginBG19,DBLP:conf/emnlp/BoginGB19,DBLP:conf/emnlp/YuYYZWLR19,DBLP:journals/corr/abs-2004-03125} and 
joint text-DB linking~\cite{DBLP:conf/acl/GuoZGXLLZ19,Wang2019RATSQLRS}, \Ours encodes the 
tagged 
sequence with BERT and lightweight subsequent layers -- two single-layer bi-directional LSTMs~\cite{DBLP:journals/neco/HochreiterS97}. Each schema component (table or field) is simply represented using the hidden state corresponding to its special token in the 
hybrid sequence. 
To better align the schema components with the question, 
\Ours augments the hybrid sequence with \emph{anchor texts}, which are automatically extracted DB cell values mentioned in the question. 
Anchor texts are appended to their corresponding fields in the hybrid sequence (Figure~\ref{fig:encoder}). The text-DB alignment is then implicitly achieved via fine-tuned BERT attention between overlapped lexical tokens.

Combined with a pointer-generator decoder~\cite{DBLP:conf/acl/SeeLM17} and \emph{schema-consistency driven search space pruning}, 
\Ours 
achieves performances comparable to or better than the state-of-the-art on the Spider (71.1\% dev, 67.5\% test with ensemble model) and WikiSQL (92.6\% dev, 91.9\% test) benchmarks, 
outperforming 
most of lately proposed models with 
task-specific architectures.\footnote{An earlier version of this model is implemented within the Photon system demonstration \url{https://naturalsql.com}~\cite{DBLP:conf/acl/ZengLHSXLK20}, with up to one anchor text per field and a less accurate anchor text matching algorithm.}
Through in-depth model comparison and error analysis, we show the proposed architecture is effective for generalizing over natural language variations and memorizing structural patterns, but struggles in compositional generalization and suffers from lack of explainability. This leads us to conclude that cross-domain text-to-SQL still poses many unsolved challenges, requiring models to demonstrate generalization over both natural language variation and structure composition while training data is often sparse.

\hide{
However, neural architectures accommodating the graph structures of relational DB schema and SQL syntax often require additional data processing and more computation resources.
We seek to explore the potential of sequence-to-sequence modeling on this task, motivated by its easy-to-adapt representation scheme and computational efficiency.
We present \Ours, a sequence-to-sequence architecture that leverages BERT to jointly represent the input question and a novel value-aware DB schema encoding.
Combined with search space pruning 
during inference, \Ours achieves state-of-the-art performance (67.2\% dev, XX\% test) on the well studied 
Spider benchmark, significantly outperforming 
models of comparable size that adopt sophisticated structure modeling.
}
\section{Model}
\label{sec:model}

In this section, we present the \Ours model 
that combines a BERT-based encoder with a sequential pointer-generator to perform end-to-end cross-DB text-to-SQL semantic parsing.

\subsection{Problem Definition}
\label{sec:picklist}
We formally defined the cross-DB text-to-SQL task as the following. Given a natural language question $\question$ and the schema $\schema = \langle \sTableSet, \sColumnSet \rangle$ for a
relational database, the parser needs to generate the corresponding SQL query $\sql$.
The schema consists of tables $\sTableSet = \left\{ \sTable_1, \dots, \sTable_N \right\}$ and fields $\sColumnSet = \{ \sColumn_{11}, \dots, \sColumn_{1|T_1|}, \dots, \sColumn_{n1}, \dots, \sColumn_{N|T_N|} \}$. 
Each table~$\sTable_i$ and each field~$\sColumn_{ij}$ has a textual name.
Some fields are \emph{primary keys}, used for uniquely indexing eachEar data record,
and some are \emph{foreign keys}, used to reference a primary key in a different table. 
In addition, each field has a \emph{data type}, $\type \in \{ \texttt{number},\, \texttt{text},\, \texttt{time},\, \texttt{boolean},\, etc. \}$.

Most existing solutions for this task do not consider DB content~\cite{DBLP:journals/corr/abs-1709-00103,DBLP:conf/emnlp/YuYYZWLR18}. Recent approaches 
show accessing DB content can significantly improve system performance~\cite{DBLP:conf/nips/LiangNBLL18,Wang2019RATSQLRS,DBLP:journals/corr/abs-2005-08314}. We consider the setting adopted by~\citet{Wang2019RATSQLRS}, where the model has access to the value set of each field instead of full DB content. For example, the field \<Property\_Type\_Code> in Figure~\ref{fig:encoder} can take one of the following values: $\{$``Apartment'', ``Field'', ``House'', ``Shop'', ``Other''$\}$. We call such value sets \emph{picklists}. 
This setting protects individual data record and sensitive fields such as user IDs or credit numbers can be hidden. 

\subsection{Question-Schema Serialization and Encoding}
\label{sec:vase_encoder}


As shown in Figure~\ref{fig:encoder}, we represent each table with its table name followed by its fields. Each table name is preceded by the special token \<[T]> and each field name is preceded by 
\<[C]>. The representations of multiple tables are concatenated to form a serialization of the schema, which is surrounded by two \<[SEP]> tokens 
and concatenated to the question. Finally, following the input format of BERT, the question is preceded by \<[CLS]> to form the hybrid question-schema serialization 
\begin{align*}
    X = &\texttt{[CLS]}, \question, \texttt{[SEP]},\texttt{[T]}, t_1, \texttt{[C]}, c_{11} \dots, c_{1|T_1|}, \\
    &\texttt{[T]}, t_2, \texttt{[C]}, c_{21}, \dots, \texttt{[C]}, c_{N|T_N|}, \texttt{[SEP]}.
\end{align*}

$X$ is encoded with 
BERT, followed by a bi-directional LSTM 
to form the base 
encoding $\vect{h}_{\text{X}}\in\Real^{|X|\times n}$. The question segment of $\vect{h}_{\text{X}}$ is passed through another bi-LSTM to obtain the question encoding $\vect{h}_{\text{Q}}\in\Real^{|Q|\times n}$. 
Each table/field is represented using the slice of $\vect{h}_{\text{X}}$ corresponding to its special token \<[T]>/\<[C]>.

\paragraph{Meta-data Features} We train dense look-up features to represent meta-data of the schema. This includes whether a field is a primary key ($\vect{f}_{\text{pri}}\in\Real^{2\times n}$), whether the field appears in a foreign key pair ($\vect{f}_{\text{for}}\in\Real^{2\times n}$) and the data type of the field ($\vect{f}_{\text{type}}\in\Real^{|\type|\times n}$). These meta-data features are fused with the base encoding of the schema component 
via a feed-forward layer $g$ ($\Real^{4n}\rightarrow\Real^{n}$) to obtain the following encoding output: 
\begin{gather}
    \begin{align}
        &\vect{h}_S^{t_i} = g([\vect{h}_{\text{X}}^p; \vect{0}; \vect{0}; \vect{0}]), \\
        &\vect{h}_S^{c_{ij}} = g([\vect{h}_{\text{X}}^q; \vect{f}_{\text{pri}}^u; \vect{f}_{\text{for}}^v; \vect{f}_{\text{type}}^w]) \\
        &\ \ \ \ \ \ = \text{ReLU}(\vect{W}_g[\vect{h}_{\text{X}}^m; \vect{f}_{\text{pri}}^u; \vect{f}_{\text{for}}^v; \vect{f}_{\text{type}}^w] + \vect{b}_g) \nonumber \\
        &\vect{h}_S = [\vect{h}^{t_1}, \dots, \vect{h}^{t_{|\sTableSet|}}, \vect{h}^{c_{11}}, \dots, \vect{h}^{c_{N|T_N|}}] \in\Real^{|\schema|\times n},
    \end{align}    
\end{gather}
where $p$ is the index of 
\<[T]> associated with table $t_i$ in $X$ and $q$ is the index of 
\<[C]> associated with field $c_{ij}$ in X. $u$, $v$ and $w$ are feature indices indicating the properties of $\sColumn_{ij}$. $[\vect{h}_{\text{X}}^m; \vect{f}_{\text{pri}}^u; \vect{f}_{\text{for}}^v; \vect{f}_{\text{type}}^w]\in\Real^{4n}$ is the concatenation of the four vectors. The meta-data features 
are specific to fields and the table representations are fused with place-holder $\vect{0}$ vectors.

\subsection{Bridging}
\label{sec:bridge}
\begin{figure*}[t]
	\centering
	\includegraphics[width=1.03\textwidth]{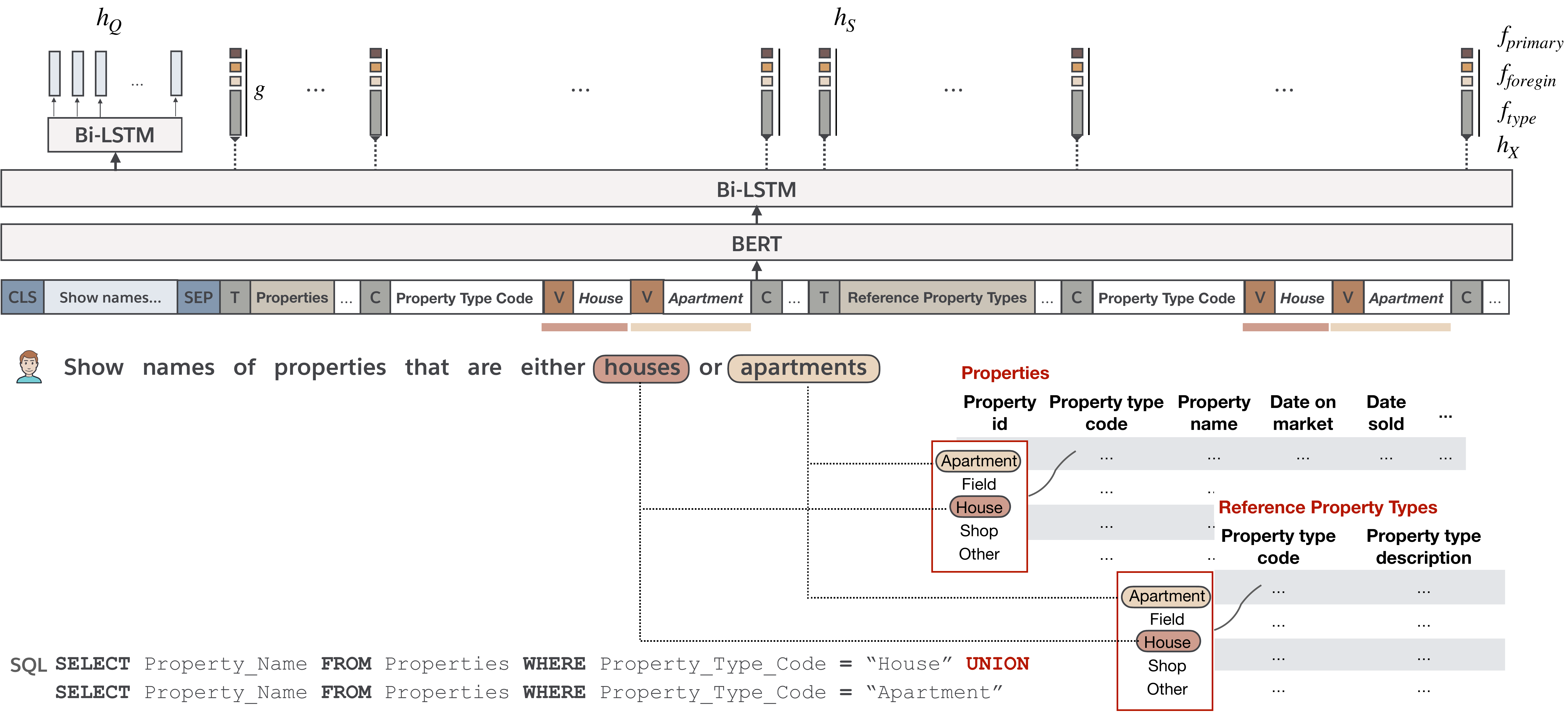}
	\caption{The \Ours encoder. The two phrases ``houses'' and ``apartments'' in the input question both matched to two DB fields. 
	The matched 
	values are appended to the corresponding field names in the hybrid sequence.} 
	\label{fig:encoder}
\end{figure*}

Modeling only the table/field names and their relations is not always enough to capture the semantics of the schema and its dependencies with the question.
Consider the example in Figure~\ref{fig:encoder}, \<Property\_Type\_Code> is a general expression not explicitly mentioned in the question, and without access to the set of possible field values, 
it is difficult to associate ``houses'' and ``apartments'' with it. 
To resolve this problem, 
we make use of \emph{anchor text} to link value mentions in the question with the corresponding DB fields. 
We perform fuzzy string match between $\question$ and the picklist of each field in the DB. The matched field values (anchor texts) are inserted into the question-schema representation $X$, succeeding the corresponding field names and separated by the special token \<[V]>. If multiple values were matched for one field, we concatenate all of them in matching order 
(Figure~\ref{fig:encoder}). 
If a question mention is matched with values in multiple fields. We add all matches and let the model learn to resolve ambiguity.\footnote{This approach may over-match anchor texts from fields other than those appeared in the correct SQL query, but keeping the additional matches in $X$ may provide useful signal rather than noise.} 

The anchor texts provide additional lexical clues for BERT to identify the corresponding mention in $\question$. And we name this mechanism ``bridging''.

\subsection{Decoder}
We use an LSTM-based pointer-generator~\cite{DBLP:conf/acl/SeeLM17} with multi-head attention~\cite{DBLP:conf/nips/VaswaniSPUJGKP17} as the decoder. The decoder is initiated with the final state of the question encoder. At each step, the decoder performs one of the following actions: generating a token from the vocabulary $\sVocab$, copying a token from the question $\question$ or copying a schema component from $\schema$.

Mathematically, at each step $t$, given the decoder state $\vect{s}_t$ and the encoder representation $[\vect{h}_Q; \vect{h}_S]\in\Real^{(|\question|+|\schema|)\times n}$, we compute the multi-head attention as defined in~\citet{DBLP:conf/nips/VaswaniSPUJGKP17}:
\begin{gather}
    \small
    \begin{align}
        e_{tj}^{(h)} &= \frac{\vect{s_t} W_U^{(h)} (\vect{h_j} W_V^{(h)})^\top}{\sqrt{n / H}}
            ;\quad
            \alpha_{tj}^{(h)} = \softmax_j \left\{ e_{tj}^{(h)} \right\}
            \\
            \vect{z}_t^{(h)} &= \sum_{j=1}^{|\question|+|\schema|} \alpha_{tj}^{(h)} (\vect{h_j} W_V^{(h)})
            ;\quad
            \vect{z}_t = \bigl[\vect{z}_t^{(1)}; \cdots; \vect{z}_t^{(H)}\bigr],
    \end{align}
\end{gather}
where $h\in[1,\hdots,H]$ is the head number and $H$ is the total number of heads. 

The probability of generating from $\sVocab$ and the 
output distribution is defined as
\begin{align}
    & \pgen^t = \text{sigmoid}(\vect{s}_t\vect{W}_{\text{gen}}^s + \vect{z}_t\vect{W}_{\text{gen}}^z + \vect{b}_{\text{gen}}) \\
    & \pout^t = \pgen^t P_{\sVocab}(y_t) + (1 - \pgen^t)\sum_{j:\Tilde{X}_j=y_t}\alpha_{tj}^{(H)},
\end{align}
where $P_{\sVocab}(y_t)$ is the softmax LSTM output distribution and $\Tilde{X}$ is the length-($|\question|+|\schema|$) 
sequence that consists of only the question words and special tokens \<[T]> and \<[C]> from $X$. We use the attention weights of the last head to compute the pointing distribution\footnote{In practice we find this approach better than using just one head or the average of multiple head weights (\S\ref{sec:performance-attn-heads}).}.

We extend the input state to the LSTM decoder using the \emph{selective read} 
proposed by~\citet{DBLP:conf/acl/GuLLL16}. 
The technical details of this extension can be found in \S\ref{sec:selective_read}.

\subsection{Schema-Consistency Guided Decoding}
\label{sec:schema-consistency}
\hide{
The search space of a sequence decoder is exponential to the generation vocabulary size plus $|\question|+|\schema|$. 
Previous work have made use of the SQL syntax 
and proposed abstract-syntax tree based decoders which guarantees the output SQL is grammatically correct~\cite{DBLP:conf/acl/GuoZGXLLZ19,Wang2019RATSQLRS,DBLP:journals/corr/abs-2005-08314}. 
Compared to sequence-decoders, tree-based decoders are typically less efficient both speed- and memory-wise.
}

We 
propose simple heuristics for pruning the search space of the sequence decoders, 
based on SQL syntax constraints and the fact that the DB fields appeared in each SQL clause must only come from the tables in the \<FROM> clause. 

\paragraph{Generating SQL Clauses in Execution Order} We rearrange the clauses of each SQL query in the 
training set into the standard DB execution order~\cite{DBLP:books/daglib/0078665} shown in table~\ref{tab:clause_order}.
For example, the SQL \<SELECT COUNT(*) FROM Properties> is converted\footnote{More complex examples can be found in Table~\ref{tab:execution_order}.} to \<FROM Properties SELECT COUNT(*)>. 
\hide{The execution order represents a coarse-to-fine data processing flow which first selects and joins relevant tables, specifies data filtering criteria, selects relevant fields and finally performs sorting and offsetting on the selected data. We hypothesize that generating SQL query in this order enables the model to learn this 
pattern and biases the model towards the subspace with higher schema consistency. 
Moreover, }

\begin{table}[t]
    \setlength{\tabcolsep}{1.4pt}
    \centering
    \scalebox{0.84}{
    \begin{tabular}{c c c c c c c c}
    \toprule
    \underline{Written:} & \<SELECT> & \<FROM> & \<WHERE> & \<GROUPBY> & \<HAVING> & \<ORDERBY> & \<LIMIT> \\
    \underline{Exec:} & \<FROM> & \<WHERE> & \<GROUPBY> & \<HAVING> & \<SELECT> & \<ORDERBY> & \<LIMIT> \\
    \bottomrule
    \end{tabular}}
    \caption{The written order vs. execution order of all SQL clauses appeared in Spider.}
    \label{tab:clause_order}
\end{table}

We can show that a SQL query 
with clauses in execution order satisfies the following lemma. 
\begin{lemma}
\label{lemma:field_ordering}
Let $Y_{\text{exec}}$ be a SQL query with clauses arranged in execution order, 
then any table field in $Y_{\text{exec}}$ must appear after the table.
\end{lemma}

As a result, we adopt a binary attention mask $\xi$ 
\begin{equation}
    \Tilde{\alpha}_{t}^{(H)} = \alpha_{t}^{(H)} \cdot \xi
\end{equation}
which initially has entries corresponding to all fields set to 0. Once a table $t_i$ is decoded, we set all entries in $\xi$ corresponding to \{$c_{i1}, \dots, c_{i|T_i|}$\} to 1. 
This allows the decoder to only search in the space specified by the condition in Lemma~\ref{lemma:field_ordering} with little overhead in decoding speed. 

In addition, we observe that a valid SQL query satisfies the following token transition lemma.
\begin{lemma}
\label{lemma:token_neighbor}
\textbf{Token Transition:} Let $Y$ be a valid SQL query, then any table/field token in $Y$ can only appear after a SQL reserved token; any value token in $Y$ can only appear after a SQL reserved token or a value token.
\end{lemma}
We use this heuristics to prune the set of candidate tokens at each decoding step. It is implemented via vocabulary masking.
\section{Related Work}
\label{sec:related_work}

\paragraph{Text-to-SQL Semantic Parsing}
Recently the field has witnessed a re-surge of interest for text-to-SQL semantic parsing~\cite{androutsopoulos_ritchie_thanisch_1995}, by virtue of newly released large-scale 
datasets~\cite{DBLP:journals/corr/abs-1709-00103,DBLP:conf/emnlp/YuYYZWLR18,DBLP:conf/emnlp/YuYYZWLR19} and matured neural network modeling tools~\cite{DBLP:conf/nips/VaswaniSPUJGKP17,DBLP:conf/naacl/ShawUV18,DBLP:conf/naacl/DevlinCLT19}. While existing models have surpassed human performance on benchmarks consisting of single-table and simple SQL queries~\cite{DBLP:journals/corr/abs-1902-01069,lyu2020hybrid,DBLP:journals/corr/abs-1908-08113}, ample space of improvement still remains for the Spider benchmark\footnote{\url{https://yale-lily.github.io/spider}} which consists of relational DBs and complex SQL queries. 

Recent architectures proposed for this problem show increasing complexity in both the encoder and the decoder~\cite{DBLP:conf/acl/GuoZGXLLZ19,Wang2019RATSQLRS,DBLP:journals/corr/abs-2004-03125,DBLP:journals/corr/abs-2007-08970}. \citet{DBLP:conf/acl/BoginBG19,DBLP:conf/emnlp/BoginGB19} proposed to encode relational DB schema as a graph and also use the graph structure to guide decoding. \citet{DBLP:conf/acl/GuoZGXLLZ19} proposes schema-linking and SemQL, an intermediate SQL representation customized for questions in the Spider dataset which was synthesized via a tree-based decoder. 
\citet{Wang2019RATSQLRS} proposes RAT-SQL, a unified graph encoding mechanism which effectively covers relations in the schema graph and its linking with the question. 
The overall architecture of RAT-SQL is deep, consisting of 8 relational self-attention layers~\cite{DBLP:conf/naacl/ShawUV18} on top of BERT-large.
In comparison, \Ours uses BERT combined with minimal subsequent layers. It uses a simple sequence decoder with search space-pruning heuristics and applies little abstraction to the SQL surface form.

\paragraph{Seq2Seq Models for Text-to-SQL Semantic Parsing} Many work have applied sequence-to-sequence models to solve semantic parsing, treating it as a translation problem~\cite{DBLP:conf/acl/DongL16,DBLP:conf/lrec/LinWZE18}. 
Text-to-SQL models take both the natural language question and the DB as input, and a commonly used input representation in existing work is to concatenate the question with a squential version of the DB schema (or table header if there is only a single table).~\citet{DBLP:journals/corr/abs-1709-00103} proposed the Seq2SQL model which first adopted this representation and tested it on WikiSQL.~\citet{DBLP:journals/corr/abs-1902-01069} first demonstrated that encoding such representation with BERT can achieve upperbound performance on the WikiSQL benchmark. 
Our work shows that such sequence representation encoded with BERT is also effective for synthesizing complex SQL queries issued to multi-table databases. Concurrently, \citet{Suhr2020} adopted a transformer model with BERT as encoder on Spider; \citet{DBLP:journals/corr/abs-2010-12725} shows that the T5 model~\cite{DBLP:journals/jmlr/RaffelSRLNMZLL20} with 3 billion parameters achieves the state-of-the-art performance on Spider. However, both of these two models do not use DB content. In addition, \Ours achieves comparable performance with a significantly smaller model. Especially, the \Ours decoder is a single-layer LSTM compared to the 12-layer transformer in T5.

\paragraph{Text-to-SQL Semantic Parsing with DB Content} 
\citet{DBLP:conf/emnlp/YavuzG0Y18} uses question-value matches to achieve high-precision condition predictions on WikiSQL. \citet{DBLP:conf/acl/ShawMCPA19} also shows that value information is critical to the cross-DB semantic parsing tasks, yet the paper reported negative results augmenting an GNN encoder with BERT and the overall model performance is much below state-of-the-art.
While previous work such as~\citet{DBLP:conf/acl/GuoZGXLLZ19,Wang2019RATSQLRS,DBLP:journals/corr/abs-2005-08314} use feature embeddings or relational attention layers 
to explicitly model schema linking, \Ours models the linking implicitly with BERT and lexical anchors. 

In addition, instead of directly taking DB content as input, some models leverage the content by training the model with SQL query execution~\cite{DBLP:journals/corr/abs-1709-00103} or performing execution-guided decoding during inference~\cite{DBLP:journals/corr/abs-1807-03100}. To our best knowledge, such methods have been tested exclusively on the WikiSQL benchmark. 

\paragraph{Joint Representation of Textual-Tabular Data and Pre-training} 

\Ours is a general framework for jointly representing question, DB schema and the relevant DB cells. It has the potential to be applied to a wider range of problems that requires joint textual-tabular data understanding. Recently,~\citet{DBLP:journals/corr/abs-2005-08314} proposes TaBERT, an LM for jointly representing textual and tabular data pre-trained over millions of web tables. Similarly,~\citet{tapas} proposes \textsc{TaPas}, a pretrained text-table LM that supports arithmetic operations for weakly supervised table QA. Both TaBERT and \textsc{TaPaS} focus on contextualizing text with a single table. TaBERT was applied to Spider by encoding each table individually and modeling cross-table correlation through hierarchical attention. In comparison, \Ours serialized the relational DB schema and uses BERT to model cross-table dependencies. \tabert adopts the ``content snapshot'' mechanism which retrieves 
table rows most similar to the input question and jointly encodes them with the table header. Compared to \Ours which uses the anchor texts, table rows are not always available if DB content access is restricted. Furthermore, anchor texts provide more focused signals that link the text and the DB schema. 

\section{Experiment Setup}
\label{sec:experiments}
\subsection{Dataset}
\label{sec:dataset}
We evaluate \Ours using two well-studied cross-database text-to-SQL benchmark datasets: Spider~\cite{DBLP:conf/emnlp/YuYYZWLR18} and WikiSQL~\cite{DBLP:journals/corr/abs-1709-00103}. 
Table~\ref{tab:data_stats} shows the statistics of the train/dev/test splits of the datasets. In the Spider benchmark, the train/dev/test databases do not overlap, and the test set is hidden from public. For WikiSQL, 49.6\% of its dev tables and 45.1\% of its test tables are not found in the train set. Therefore, both datasets necessitates the ability of models to generalize to unseen schema. 

We run hyperparameter search and analysis on the dev set and report the test set performance only using our best approach.
\begin{table}[t]
    \centering
    \scalebox{0.84}{
    \begin{tabular}{c|ccc|cc}
    \toprule
         & \multicolumn{3}{c|}{\bfseries Spider} & \multicolumn{2}{c}{\bfseries WikiSQL} \\
         &  \# Q & \# SQL & \#DB  & \# Q & \# Table \\
    \midrule 
        Train & 8,695 & 4,730 & 140 & 56,355 & 17,984 \\
        Dev & 1,034 & 564 & 20 & 8,421 & 1,621 \\
        Test & 2,147 & -- & 40 & 15,878 & 2,787\\
    \bottomrule
    \end{tabular}}
    \caption{Text-to-SQL Dataset Statistics}
    \label{tab:data_stats}
\end{table}

\subsection{Evaluation Metrics}
We report the official evaluation metrics proposed by the Spider and WikiSQL authors. 

\paragraph{Exact Match (EM)} This metrics checks if the predicted SQL exactly matches the ground truth SQL. It is a performance lower bound as a semantically correct SQL query may differ from the ground truth SQL query in surface form.

\paragraph{Exact Set Match (E-SM)} This metrics evaluates the structural correctness of the predicted SQL by checking the orderless set match of each SQL clause in the predicted query w.r.t. the ground truth. It ignores errors in the predicted values.

\paragraph{Execution Accuracy (EA)} This metrics checks if the predicted SQL is executable on the target DB and if the execution results of match those of the ground truth. It is a performance upper bound as two SQL queries with different semantics can execute to the same results on a DB.

\subsection{Implementation Details}
\label{sec:implementation_details}
\begin{figure}[t]
    \centering
    \includegraphics[width=0.42\textwidth]{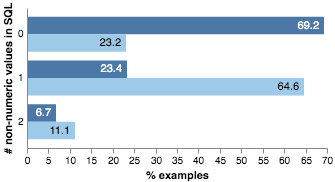}
    \caption{Distribution of \# non-numeric values in the ground truth SQL queries in the \colbox{cspiderback}{\textcolor{cspider}{Spider}} and \colbox{cwikisqlback}{\textcolor{cwikisql}{WikiSQL}} dev sets.}
    \label{fig:anchor_text_hist}
\end{figure} 
\paragraph{Anchor Text Selection} Given a DB, we compute the pickist of each field using the official DB files. We developed a fuzzy matching algorithm to match a question to possible value mentions in the DB (described in detail in \S\ref{sec:picklist-eval}). We include up to $k$ matches per field, 
and break ties by taking the longer match. 
We exclude all number matches as a number mention in the question may not correspond to a DB cell (e.g. it could be a hypothetical threshold as in ``shoes lower than $\$50$'') or cannot effectively discriminate between different fields.

Figure~\ref{fig:anchor_text_hist} shows the distribution of non-numeric values in the ground truth SQL queries from the Spider and WikiSQL dev sets. For Spider, 31\% of the examples contain one or more non-numeric values in the ground truth queries and can potentially benefit from the bridging mechanism. For WikiSQL the ratio is significantly higher, with 76.8\% of the ground truth SQL queries contain one or more non-numeric values. On both datasets, the proportion of ground truth SQL queries containing $>2$ non-numeric values are negligible (0.8\% for Spider and 1.1\% for WikiSQL). Based on this analysis, we set $k=2$ in all our experiments.

\hide{We convert the question and field values into lower cased character sequences and compute the longest sub-sequence match with heuristically determined matching boundaries\footnote{This allows us to capture plural/singular nouns and other trivial string variations.}. We define 
a matching score function and only consider matches that score above a tuned threshold 
}

\paragraph{Data Repair} The original Spider dataset contains errors in both the example files and database files. We manually corrected some errors in the train and dev examples. For comparison with other models in \S\ref{res:main}, we report metrics using the official dev/test sets. For our own ablation study and analysis, we report metrics using the corrected dev files. We also use a high-precision heuristics to identify missing foreign key pairs in the databases and combine them with the released ones during training and inference: if two fields of different tables have identical name and one of them is a primary key, we count them as a foreign key pair\footnote{We exclude common field names such as ``name'', ``id'' and ``code'' in this procedure.}.

\paragraph{Training} We train our model using cross-entropy loss. We use Adam-SGD~\cite{DBLP:journals/corr/KingmaB14} with default parameters and a mini-batch size of 32. We use the uncased BERT-large model from the Huggingface's transformer library~\cite{Wolf2019HuggingFacesTS}. We set all LSTMs to 1-layer and use 8-head attention between the encoder and decoder. 

\begin{itemize}
    \item Spider: We set the LSTM hidden layer dimension to 400. We train a maximum of 100k steps. We set the learning rate to $5e^{-4}$ in the first 5,000 iterations and shrink it to 0 with the L-inv function (\S\ref{sec:linv}). We fine-tune BERT with a fine-tuning rate linearly increasing from $3e^{-5}$ to $6e^{-5}$ in the first 4,000 iterations and decaying to 0 according to the L-inv function. We randomly permute the table order in a DB schema and drop one table which does not appear in the ground truth with probability 0.3 in every training step. 
    The training time of our model on an NVIDIA A100 GPU is approximately 51.5h (including intermediate results verification time). 
    \vspace{-2mm}
    \item WikiSQL: We set the LSTM hidden layer dimension to 512. We train a maximum of 50k steps and set the learning rate to $5e^{-4}$ in the first 4,000 iterations and shrink it to 0 with the L-inv function. We fine-tune BERT with a fine-tuning rate linearly increasing from $3e^{-5}$ to $6e^{-5}$ in the first 4,000 iterations and decaying to 0 according to the L-inv function. The training time of our model on an NVIDIA A100 GPU is approximately 6h (including intermediate results verification time).
\end{itemize}

\paragraph{Decoding} The decoder uses a generation vocabulary consisting of 70 SQL keywords and reserved tokens, plus the 10 digits 
to generate numbers not explicitly mentioned in the question (e.g. ``first'', ''second'', ``youngest'' etc.). We use a beam size of 64 
for leaderboard evaluation. All other experiments uses a beam size of 16.
We use schema-consistency guided decoding during inference only. It cannot guarantee schema consistency\footnote{Consider the example SQL query shown in Table~\ref{tab:lemma_1_exception} which satisfies the condition of Lemma~\ref{lemma:field_ordering}, the table \<VOTING\_RECORD> only appears in the first sub-query, and the field \<VOTING\_RECORD.PRESIDENT\_Vote> in the second sub-query is out of scope.} and we run a static SQL correctness check on the beam search output to eliminate predictions that are either syntactically incorrect or violates schema consistency\footnote{Prior work such as~\citep{DBLP:journals/corr/abs-1807-03100} performs the more aggressive execution-guided decoding. However, it is difficult to apply this approach 
to complex SQL queries~\cite{DBLP:journals/corr/abs-1709-00103}. We build a static SQL analyzer on top of the Mozilla SQL Parser (\url{https://github.com/mozilla/moz-sql-parser}). 
Our static checking approach handles complex SQL queries and avoids DB execution overhead.}. 
For WikiSQL, the static check also makes sure that the output query conforms to the SQL sketch used to create the dataset~\cite{DBLP:journals/corr/abs-1709-00103}.
If no predictions in the beam satisfy the two criteria, we output a default SQL query which count the number of entries in the first table.
\section{Results}
\subsection{End-to-end Performance Evaluation}
\label{res:main}

\begin{table}[t]
    \setlength{\tabcolsep}{4pt}
    \centering
    \scalebox{0.84}{
    \begin{tabular}{lrr}
        \toprule
        \bfseries Model & \bfseries Dev & \bfseries Test \\
        \midrule
        Global-GNN \citep{DBLP:conf/emnlp/BoginGB19} $\spadesuit$ & 52.7 & 47.4 \\
        EditSQL + BERT \citep{DBLP:conf/emnlp/YuYYZWLR19} & 57.6 & 53.4 \\
        GNN + Bertrand-DR \citep{kelkar2020bertrand} & 57.9 & 54.6 \\
        IRNet + BERT \citep{DBLP:conf/acl/GuoZGXLLZ19} & 61.9 & 54.7 \\
        RAT-SQL v2 $\spadesuit$ \citep{Wang2019RATSQLRS} & 62.7 & 57.2 \\
        RYANSQL + BERT$_{\text{L}}$ \citep{DBLP:journals/corr/abs-2004-03125} & 66.6 & 58.2 \\
        SmBoP + BART \citep{DBLP:journals/corr/abs-2010-12412} & 66.0 & 60.5 \\
        RYANSQL v2 + BERT$_{\text{L}}$ $\diamond$ & 70.6 & 60.6 \\
        RAT-SQL v3 + BERT$_{\text{L}}$ $\spadesuit$ \citep{Wang2019RATSQLRS}& \emph{69.7} & \emph{65.6} \\
        \midrule
        \Ours v1 $\spadesuit$ $\heartsuit$ \citep{DBLP:conf/emnlp/LinSX20} & 65.5 & 59.2 \\
        \Ours$_{\text{L}}$ (ours) $\spadesuit$ $\heartsuit$ & 70.0 & 65.0 \\
        \Ours$_{\text{L}}$ (ours, ensemble) $\spadesuit$ $\heartsuit$ & 71.1 & 67.5 \\
        \bottomrule
        \toprule
        \bfseries Model & \bfseries Dev & \bfseries Test \\
        \midrule
        %
        AuxNet + BART $\diamond$ $\spadesuit$ $\heartsuit$ & - & 62.6 \\
        \midrule
        \Ours v1 $\spadesuit$ $\heartsuit$ \citep{DBLP:conf/emnlp/LinSX20} & 65.3 & 59.9 \\
        \Ours$_{\text{L}}$ (ours) $\spadesuit$ $\heartsuit$ & 68.0 & 64.3 \\
        \Ours$_{\text{L}}$ (ours, ensemble) $\spadesuit$ $\heartsuit$ & 70.3 & 68.3 \\
        \bottomrule
    \end{tabular}}
    
    \caption{Exact set match (top) and execution accuracy (bottom) on the Spider dev and test sets, compared to the other top-performing approaches on the leaderboard as of Dec 20, 2020. The test set results were issued by the Spider team. BERT$_\text{L}$ denotes BERT$_{\text{LARGE}}$. 
        $\diamond$ denotes approaches without reference in publication. $\spadesuit$ denotes approaches using DB content. $\heartsuit$ denote approaches generating executable output.} 
    \label{tab:main-leaderboard}
    \vspace{-1\baselineskip}
\end{table}

\subsubsection{Spider}
Table~\ref{tab:main-leaderboard} shows the 
performance of \Ours compared to other approaches ranking at the top of the Spider leaderboard. 
\Ours v1 is our model described in the original version of the paper. Comparing to \Ours v1, the current model is trained with BERT-large with an improved anchor text matching algorithm (\S\ref{sec:picklist-eval}).
\Ours$_\text{L}$ performs very competitively, significantly outperforming most of recently proposed architectures with more complicated, task-specific layers (Global-GNN, EditSQL+BERT, IRNet+BERT, RAT-SQL v2, RYANSQL+BERT$_\text{L}$). 
It also performs better than or comparable to models that explicitly model compositionality in the decoder (SmBoP, RAT-SQL v3L+BERT$_\text{L}$, RYANSQL).\footnote{Simply comparing the leaderboard performances does not allow precise gauging of different modeling trade-offs, as all leaderboard entries adopt some customized pre- and post- processing of the data. For example, the schema-consistency guided decoding adopted by \Ours is complementary to other models. \Ours synthesizes a complete SQL query while several other models do not synthesize values and synthesize the \<FROM> clause in a post-processing step~\cite{Wang2019RATSQLRS}.} 
In addition, \Ours generates executable SQL queries by copying values from the input question while most existing models only predicts the SQL syntax skeleton.\footnote{We believe the execution accuracy can be further improved by having the model copying the anchor texts and plan to explore this in future work.} As of Dec 20, 2020, \Ours ranks top-1 on the Spider leaderboard by execution accuracy.


\begin{table}[t]
    \setlength{\tabcolsep}{2.2pt}
    \centering
    \scalebox{0.83}{
    \begin{tabular}{lrrrrr}
        \toprule
        \bfseries Model  & \bfseries Easy & \bfseries Medium & \bfseries Hard & \bfseries Ex-Hard & \bfseries All \\
        \midrule
        count & 250 & 440 & 174 & 170 & 1034 \\
        \midrule
        \textit{Dev} \\
        RAT-SQL v3+B$_L$ $\spadesuit$ &  86.4 & \textbf{73.6} & \textbf{62.1} & 42.9 & 69.7 \\
        \Ours$_\text{L}$ $\spadesuit$ & \textbf{89.1} & 72.2 & 56.3 & \textbf{50.0} & \textbf{70.0} \\
        \hdashline
        \Ours$_\text{L,ens}$ $\spadesuit$ & 89.1 & 71.7 & 62.1 & 51.8 & 71.1 \\
        \midrule
        \textit{Test} \\
        IRNet+B &  77.2 & 58.7 & 48.1 & 25.3 & 54.7 \\
        \Ours$_\text{L}$ $\spadesuit$ & \textbf{85.1} & 71.2 & 55.3 & 36.1 & 65.0 \\
        RAT-SQL v3+B$_L$ $\spadesuit$ &  83.0 & \textbf{71.3} & \textbf{58.3} & \textbf{38.4} & \textbf{65.6} \\
        \hdashline
        \Ours$_\text{L,ens}$ $\spadesuit$ & 85.3 & 73.4 & 59.6 & 40.3 & 67.5 \\
        \bottomrule
    \end{tabular}}
    \caption{E-SM by SQL hardness level compared to other approaches on Spider leaderboard.} 
    \label{tab:difficulty_compare}
\end{table}

RAT-SQL v3+BERT$_\text{L}$ outperforms \Ours in terms of exact set match with a small margin.
We further look at the performance comparison 
between the two models across different SQL query hardness level (Table~\ref{tab:difficulty_compare}). 
Overall, \Ours outperforms RAT-SQL v3+BERT$_L$ in the easy category but underperforms it in the other three categories, with considerable gaps in medium and hard. 

We hypothesize that differences in both the encoders and decoders of the two models have contributed to the performance differences. The RAT-SQL encoder and decoder are designed with compositional inductive bias. It models the relational DB schema as a graph encoded with relational self-attention. The decoder uses SQL-syntax guided generation~\cite{DBLP:conf/acl/YinN17}. \Ours, on the other hand, adopts a Seq2Seq architecture. In addition, RAT-SQL v3 models the lexical mapping between question-schema and question-value via a graph with edge labeled by the matching condition (full-word match, partial match, etc.). 
\Ours represents 
the same information in a tagged sequence and 
uses fine-tuned BERT to implicitly obtain such mapping. While the anchor text selection algorithm (\S\ref{sec:implementation_details}) has taken into account string variations, BERT may not be able to capture the linking when string variations exist --
it has not seen tabular input during pre-training.
The tokenization scheme adopted by BERT and other pre-trained LMs (e.g. GPT-2) 
cannot effectively capture partial string matches 
in a novel input
(e.g. ``cats'' and ``cat'' are two different words in the vocabularies of BERT and GPT-2). 
Pre-training the architecture using more tables and heuristically aligned text may alleviate this problem~\cite{DBLP:journals/corr/abs-2005-08314,tapas}. 
Finally, we notice that ensembling three models (averaging the output distributions at each decoding step) trained with different random seeds improves the performance in all SQL hardness levels, especially in \emph{medium}, \emph{hard} and \emph{extra-hard}.


\subsubsection{WikiSQL}

\begin{table}[t]
    \setlength{\tabcolsep}{5pt}
    \centering
    \hspace{-10pt}
    \scalebox{0.76}{
    \begin{tabular}{lrrrr}
        \toprule
        \bfseries Model  & \multicolumn{2}{c}{\bfseries Dev} & \multicolumn{2}{c}{\bfseries Test} \\
        & \bfseries EM & \bfseries EX & \bfseries EM & \bfseries EX \\
        \midrule
        SQLova~\cite{DBLP:journals/corr/abs-1902-01069} & 81.6 & 87.2 & 80.7 & 86.2 \\
        X-SQL~\cite{he2019x} & 83.8 & 89.5 & 83.3 & 88.7 \\
        IE-SQL~\cite{ma-etal-2020-mention} & 84.6 & 88.7 & 84.6 & 88.8 \\
        NL2SQL $\spadesuit$~\cite{DBLP:journals/corr/abs-1910-07179} & 84.3 & 90.3 & 83.7 & 89.2 \\
        HydraNet~\cite{lyu2020hybrid}  & 83.6 & 89.1 & 83.8 & 89.2 \\
        \Ours$_\text{L}$ $\spadesuit$ & \textbf{86.2} & \textbf{91.7} & \textbf{85.7} & \textbf{91.1} \\
        \midrule
        SQLova${+\text{EG}}$~\cite{DBLP:journals/corr/abs-1902-01069} & 84.2 & 90.2 & 83.6 & 89.6 \\
        NL2SQL${+\text{EG}}$ $\spadesuit$~\cite{DBLP:journals/corr/abs-1910-07179} & 85.4 & 91.1 & 84.5 & 90.1 \\
        X-SQL${+\text{EG}}$~\cite{he2019x} & 86.2 & 92.3 & 86.0 & 91.8 \\
        \Ours$_\text{L}{+\text{EG}}$ $\spadesuit$ & 86.8 & \textbf{92.6} & 86.3 & 91.9 \\
        HydraNet${+\text{EG}}$~\cite{lyu2020hybrid} & 86.6 & 92.4 & 86.5 & 92.2 \\
        IE-SQL${+\text{EG}}$~\cite{ma-etal-2020-mention} & \textbf{87.9} & \textbf{92.6} & \textbf{87.8} & \textbf{92.5} \\
        \bottomrule
    \end{tabular}}
    \caption{Comparison between \Ours and other top-performing models on the WikiSQL leaderboard as of Dec 20, 2020. $\spadesuit$ denotes approaches using DB content. ${+\text{EG}}$ denotes approaches using execution-guided decoding~\cite{DBLP:journals/corr/abs-1807-03100}.} 
    \label{tab:wikisql}
\end{table}

Table~\ref{tab:wikisql} reports the comparison of \Ours$_\text{L}$ to other top-performing entries on the WikiSQL leaderboard. \Ours$_\text{L}$ achieves SOTA performance on WikiSQL, surpassing the widely cited SQLova model~\cite{DBLP:journals/corr/abs-1902-01069} by a significant margin. Among the baselines shown in Table~\ref{tab:wikisql}, SQLova is the one that's strictly comparable to \Ours as both use BERT-large-uncased.\footnote{NL2SQL uses BERT-based-uncased. Hydra-Net uses RoBERTa-Large~\cite{DBLP:conf/acl/LiuLYWCS19} and X-SQL uses MT-DNN~\cite{liu-etal-2019-multi}.} Leveraging table content (anchor texts) enables \Ours$_\text{L}$ to be the best-performing model without execution-guided (EG) decoding~\cite{DBLP:journals/corr/abs-1807-03100}. 
However, comparing to SQLova, X-SQL and HydraNet, \Ours benefits noticably less from EG. A probably reason for this is that the schema-consistency guided decoding already ruled out a significant number of SQL queries that will raise errors during execution. In addition, all models leveraging DB content during training (\Ours and NL2SQL) benefit less from EG.

\subsection{Ablation Study}
\label{sec:ablation}

\paragraph{Spider} We perform a thorough ablation study to show the contribution of each \Ours sub-component (Table~\ref{tab:ablation_difficulty}).  
The decoding search space pruning strategies we introduced (including 
schema-consistency guided decoding and static SQL correctness check) are effective, with absolute E-SM improvements 0.3\% on average. 
On the other hand, encoding techniques for jointly representing textual and tabular input contribute more. Especially, the bridging mechanism results in an absolute E-SM improvement of 1.6\%. A further comparison between \Ours with and without bridging at different SQL hardness levels 
(Table~\ref{tab:ablation_difficulty}) shows that the technique 
is especially effective at improving the model performance in the extra-hard category. We also did a fine-grained ablation study on the bridging mechanism, by inserting only the special token \<[V]> into the hybrid sequence without the anchor texts. The average model performance is not hurt and the variance decreased. This indicates that the \<[V]> tokens act as markers for columns whose value matched with the input question and contribute to a significant proportion of the performance improvement by bridging.\footnote{A similar mechanism is proposed by~\citep{DBLP:journals/corr/abs-2005-08314}, where learnable dense features are concatenated to the representations of matched utterance tokens and table/fields.} However, since the full model attained the best performance on the dev set, we keep the anchor texts in our representation.

\begin{table}
    \centering
    \scalebox{0.82}{
    \begin{tabular}{lll}
        \toprule
        \bfseries Model & \multicolumn{2}{c}{\bfseries Exact Set Match (\%)} \\
        & Mean & Max \\
        \midrule
        \Ours$_\text{L}$ & 68.2 \textpm~1.0 & 69.1 \\
        \quad - SC-guided decoding & 67.9 \textpm~0.7 (-0.3) & 69.1 (-0.0) \\
        \quad - static SQL check & 67.9 \textpm~0.6 (-0.3) & 68.8 (-0.3) \\
        \hdashline
        \quad - anchor text & 68.3 \textpm~0.4 (+0.1) & 68.8 (-0.3) \\
        \quad - table shuffle \& drop & 67.5 \textpm~1.0 \textbf{(-0.7)} & 68.7 (-0.4) \\
        \quad - meta data & 67.2 \textpm~0.2 \textbf{(-1.0)} & 67.4 \textbf{(-1.7)} \\
        \quad - bridging & 66.6 \textpm~0.5 \textbf{(-1.6)} & 67.3 \textbf{(-1.8)} \\
        \quad - BERT & 17.7 \textpm~0.7 \textbf{(-50.5)} & 18.3\textbf{(-50.8)} \\
        \bottomrule
    \end{tabular} }
\end{table}
\begin{table}
    \centering
    \scalebox{0.8}{
    \begin{tabular}{lrrrrr}
        \toprule
        \bfseries Model  & \bfseries Easy & \bfseries Medium & \bfseries Hard & \bfseries Ex-Hard & \bfseries All \\
        \midrule
        count & 250 & 440 & 174 & 170 & 1034 \\
        \midrule
        \Ours$_\text{L}$ & 85.5 & \textbf{71.5} & 56.3 & \textbf{51.8} & \textbf{69.1} \\
        \quad -bridging &  \textbf{86.3} & 70.0 & \textbf{56.9} & 42.8 & 67.3 \\
        \bottomrule
    \end{tabular}}
    \vspace{-0.4mm}
    \caption{\Ours ablations on the Spider dev set. We report the E-SM of each model variations averaged over 3 runs in the main study (top); and the E-SM of the best model in each variation in the study by SQL hardness (bottom).}
    \label{tab:ablation_difficulty}
    \vspace{-\baselineskip}
\end{table}

We also observe that the dense meta data feature encoding (\S~\ref{sec:vase_encoder}) is helpful, resulting in 1\% absolute improvement on average.
Shuffling and randomly dropping non-ground-truth tables during training also 
mildly helps our approach, as it increases the variation of DB schema seen by the model and reduces overfitting to a particular table arrangement. 
Furthermore, BERT is critical to the performance of \Ours, magnifying performance of the base model by more than three folds. This is considerably larger than the improvement prior approaches have obtained from adding BERT. Consider the performances of RAT-SQL v2 and RAT-SQL v2+BERT$_L$ in Table~\ref{tab:main-leaderboard}, the improvement using BERT$_L$ is 7\%. This shows that simply adding BERT to existing approaches results in significant redundancy in the model architecture. We perform a qualitative attention analysis in \S\ref{sec:bert_attn_vis} to show that after fine-tuning, the BERT layers effectively capture the linking between question mentions and the anchor texts, as well as the relational DB structures.


\begin{table}
    \centering
    \scalebox{0.88}{
        \begin{tabular}{lrrrr}
        \toprule
        \bfseries Model  & \multicolumn{2}{c}{\bfseries w/o EG} & \multicolumn{2}{c}{\bfseries w/ EG}\\
        & \bfseries EM & \bfseries EX & \bfseries EM & \bfseries EX \\
        \midrule
        \Ours$_\text{L}$ & \textbf{86.2} & \textbf{91.7}  & \textbf{86.8} & \textbf{92.6}\\
        \quad -bridging & 82.6 & 88.5 & 84.5 & 90.8 \\
        \bottomrule
    \end{tabular}
    }
    \caption{\Ours ablations on the WikiSQL dev set.}
    \label{tab:ablation_wikisql}
\end{table}

\paragraph{WikiSQL} The model variance on WikiSQL is much smaller than that on Spider, hence we report the ablation study results using the best model in each category. As shown in Table~\ref{tab:ablation_wikisql}, the bridging mechanism significantly enhances the model performance, especially when execution-guided decoding is not applied. As shown in Figure~\ref{fig:anchor_text_hist}, 76.8\% of the ground truth SQL queries in the WikiSQL dev set contain at least one non-numeric values. The dataset contains simple queries and the main challenge comes from interpreting filtering conditions in the \<WHERE> clause~\cite{DBLP:conf/emnlp/YavuzG0Y18}. And bridging is very effective for solving this challenge.

\hide{
Especially, the results confirm the three main contributions of \Ours. 
First, the bridging mechanism 
significantly improved the semantic grounding between the question and the DB schema by introducing the anchor text. Second, different from previous work such as RAT-SQL which develops special-purpose model layers to link the question and the DB schema, \Ours makes use of the deep attention structure of pre-trained BERT to capture such linking. As shown in the last row of Table~\ref{tab:ablation}, BERT is critical to the performance of \Ours, magnifying the performance of the ablated model by more than three folds. Finally, search space pruning is critical to sequence-based SQL generation and the schema consistency constraint we introduced is effective.
}
\subsection{Error Analysis}
\label{sec:error_analysis}

We randomly sampled 50 Spider dev set examples for which the best \Ours model failed to produce a prediction that matches the ground truth and manually categorized the errors. Each example is assigned to only the category it fits most.

\subsubsection{Manual Evaluation}
\begin{figure}[t]
    \centering
    \includegraphics[width=0.4\textwidth]{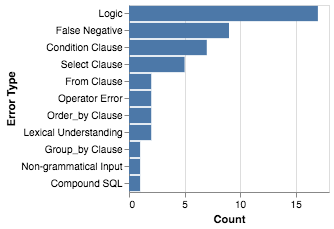}
    \caption{\Ours error type distribution (Spider dev).}
    \label{fig:error_analysis}
\end{figure}

Figure~\ref{fig:error_analysis} shows the number of examples in each category. 18\% of the examined predictions are false negatives. Among them, 5 are semantically equivalent to the ground truths; 3 use \<GROUP BY> keys different but equivalent to those of the ground truth (e.g. \<GROUY BY car\_models.name> vs. \<GROUP BY car\_models.id>); 1 has the wrong ground truth annotation. 
Among the true negatives, 
The dominant type of errors is logical mistake (18), where the output SQL query failed to represent the core logic expressed in the question. 17 have errors that can be pinpointed to specific clauses: \<WHERE> (7), \<SELECT> (5), \<FROM> (2), \<ORDER BY> (2), \<GROUP BY> (1). 2 have errors in the operators: 1 in the aggregation operator and 1 in the \<DISTINCT> operator. 1 have errors in compounding SQL clauses. 2 were due to lack of lexical and commonsense knowledge when interpreting the question, e.g. \emph{predominantly spoken language}, \emph{all address lines}. 1 example has non-grammatical natural language question.

\hide{
\paragraph{Error Causes} 
A prominent cause of errors for \Ours is irregular design and naming in the DB schema. Table~\ref{tab:error_cases} shows 3 examples where \Ours made a wrong prediction from the medium hardness level in the dev set. In the second example, the DB contains a field named ``hand'' which stores information that indicates whether a tennis player is right-handed or left-handed. While ``hand'' is already a rarely seen field name (comparing to ``name'', ``address'' etc.), the problem is worsened by the fact that the field values are acronyms which bypassed the anchor text match. Similarly, in the third example, \Ours fails to detect that ``highschooler'', normally written as ``high schooler'' is a synonym of student. Occasionally, however, \Ours still makes mistakes w.r.t. schema components explicitly mentioned in the question, as shown by the first example. Addressing such error cases could further improve its performance. 
}

\subsubsection{Qualitative Analysis}
Table~\ref{tab:error_cases} shows examples of errors from each major error type mentioned previously. 

\begin{table*}[t]
    \centering
    \scalebox{0.86}{
    \noindent\begin{tabularx}{\linewidth}{ccX}
    \toprule 
    \multirow{6}{*}{\rotatebox[origin=c]{90}{\small Logic}}
    & \parbox[c]{1em}{
      \includegraphics[width=.16in]{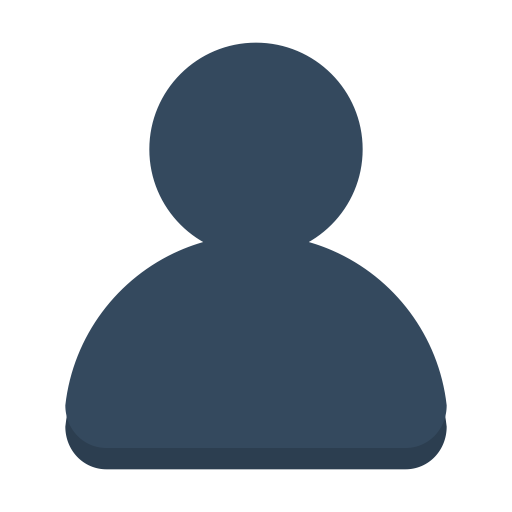}} & \emph{Find the number of concerts happened in the stadium with \underline{the highest capacity}.}  {\small\colorbox{gray}{\color{white}{\textbf{concert\_singer}}}} \\
    & \xmark & \<SELECT COUNT(*) FROM stadium JOIN concert ON stadium.Stadium\_ID = concert.Stadium\_ID ORDER BY stadium.Capacity DESC LIMIT 1> \\
    & \cmark & \<SELECT COUNT(*) FROM concert WHERE stadium\_id = (SELECT stadium\_id FROM stadium ORDER BY capacity DESC LIMIT 1)> \\
    & \parbox[c]{1em}{
      \includegraphics[width=.16in]{figures/user-alt-512.png}} & \emph{Show the names of \underline{all of the high schooler Kyle's friends}.}  {\small\colorbox{gray}{\color{white}{\textbf{network\_1}}}} \\
    & \xmark & \<SELECT Highschooler.name FROM Friend JOIN Highschooler ON Friend.friend\_id = Highschooler.ID WHERE Highschooler.name = "Kyle"> \\
    & \cmark & \<SELECT T3.name FROM Friend AS T1 JOIN Highschooler AS T2 ON T1.student\_id  = T2.id JOIN Highschooler AS T3 ON T1.friend\_id  =  T3.id WHERE T2.name  =  "Kyle"> \\
    \hdashline
    
    \multirow{3}{*}{\rotatebox[origin=c]{90}{\small Lexical Understanding}} & \parbox[c]{1em}{
      \includegraphics[width=.16in]{figures/user-alt-512.png}} & \emph{Count the number of countries for which Spanish is the \underline{predominantly} spoken language.}  {\small\colorbox{gray}{\color{white}{\textbf{world\_1}}}} \\
    & \xmark & \<SELECT COUNT(*) FROM countrylanguage WHERE countrylanguage.Language = "Spanish"> \\
    & \cmark & \<SELECT COUNT(*), MAX(Percentage) FROM countrylanguage WHERE LANGUAGE  =  "Spanish" GROUP BY CountryCode> \\
    & \parbox[c]{1em}{
      \includegraphics[width=.16in]{figures/user-alt-512.png}} & \emph{What are the full names of all \underline{left handed} players, in order of birth date?}  {\small\colorbox{gray}{\color{white}{\textbf{WTA\_1}}}} \\
    & \xmark & \<SELECT first\_name, last\_name FROM players ORDER BY birth\_date> \\
    & \cmark & \<SELECT first\_name, last\_name FROM players \underline{WHERE hand = 'L'} ORDER BY birth\_date> \\
    \hdashline
    
    & \parbox[c]{1em}{
      \includegraphics[width=.16in]{figures/user-alt-512.png}} & \emph{Which address holds the most number of students currently? List the address id and \underline{all lines}.} {\small\colorbox{gray}{\color{white}{\textbf{student\_transcripts\_tracking}}}} \\
    \multirow{3}{*}{\rotatebox[origin=c]{90}{\small Commonsense}}
    & \xmark & \<SELECT Addresses.line\_1, Students.current\_address\_id FROM Addresses JOIN Students ON Addresses.address\_id = Students.current\_address\_id GROUP BY Students.current\_address\_id ORDER BY COUNT(*) DESC LIMIT 1> \\
    & \cmark & \<SELECT Addresses.address\_id ,  Addresses.line\_1 ,  Addresses.line\_2 FROM Addresses JOIN Students ON Addresses.address\_id  =  Students.current\_address\_id GROUP BY Addresses.address\_id ORDER BY count(*) DESC LIMIT 1> \\
    \hdashline
    
    \multirow{6}{*}{\rotatebox[origin=c]{90}{\small Robustness}}
    & \parbox[c]{1em}{
      \includegraphics[width=.16in]{figures/user-alt-512.png}} & \emph{What is the model of the car with the smallest amount of \underline{horsepower}?}  {\small\colorbox{gray}{\color{white}{\textbf{car\_1}}}} \\
    & \xmark & \<SELECT cars\_data.Horsepower FROM cars\_data ORDER BY cars\_data.Horsepower LIMIT 1> \\
    & \cmark & \<SELECT T1.Model FROM CAR\_NAMES AS T1 JOIN CARS\_DATA AS T2 ON T1.MakeId = T2.Id ORDER BY T2.horsepower ASC LIMIT 1> \\
    & \parbox[c]{1em}{
      \includegraphics[width=.16in]{figures/user-alt-512.png}} & \emph{What is the \underline{total population} and \underline{average area} of countries in the continent of North America whose area is bigger than 3000?}  {\small\colorbox{gray}{\color{white}{\textbf{concert\_singer}}}} \\
    & \xmark & \<SELECT SUM(country.Population), AVG(country.Population) FROM country WHERE country.Continent = "North America" AND country.SurfaceArea > 3000$>$ \\
    & \cmark & \<SELECT SUM(country.population), AVG(country.surfacearea) FROM country WHERE country.Continent = "north america" and country.SurfaceArea > 3000$>$ \\
    \bottomrule
    
    \end{tabularx}}
    \caption{Errors cases of \Ours on the Spider dev set. The samples were randomly selected from the medium hardness level. \xmark denotes the wrong predictions made by \Ours and \cmark denotes the ground truths.}
    \label{tab:error_cases}
\end{table*}

\paragraph{Logic Errors} Logic error is a diverse category. Frequently in this case we see the model memorizing patterns seen on the training set but failed on compositional generalization. Consider the first example in this category. Superlative relation such as ``highest'' is often represented in the training set by sorting the retrieved records in descending order and taking the top 1. The model memorizes this pattern and output the correct logic for \emph{finding the stadium with the highest capacity}. It also output the correct pattern for \emph{counting the number of concerts}. Yet the correct way of combining these two logical fragments to realize the meaning in the question is to use a nested SQL query in the \<WHERE> condition. \Ours joined them flatly, and the resulting query has completely different semantics. The second example illustrates an even more interesting case. The target database is a second normal form\footnote{\url{https://en.wikipedia.org/wiki/Second_normal_form}} that triggers self-join relations (the \emph{friend} of a \emph{highschooler} is another \emph{highschooler}). Self-joins do not appear frequently in the dataset and we hypothesize it is very challenging for a Seq2Seq based model like \Ours to grasp such relation. Introducing compositional inductive bias in both the encoder and decoder could be a promising direction for solving these extra-hard cases.

\paragraph{Lexical Understanding} Another category of errors occur when the input utterance contains unseen words or phrasal expressions. While \Ours builds on top of pre-trained language models such as BERT, it is especially challenging for the model to interpret these text units grounded to the DB schema. Consider the first example in this category, ``predominantly'' means \emph{spoken by the largest percentage of the population}. It is almost impossible for the model to see such diverse natural language during supervised learning. Infusing such knowledge via pre-training is also non-trivial, but worth investigating. Continuous learning is a promising direction for this type of challenges, where the model is trained to ask clarification questions and learns new knowledge from user interaction~\cite{DBLP:conf/emnlp/YaoTYSS20}.

\paragraph{Commonsense} 
As shown by the example, US address contains two lines is a commonsense knowledge, but the model has difficulty inferring that ``all lines'' maps to ``\<line\_1> and \<line\_2>''. Again, we think continuous learning could be an effective solution for this case.

\paragraph{Robustness} The final major category of error has to do with the model blatantly ignoring information in the utterance, even when the underlying logic is not complicated, indicating that spurious correlation was captured during training~\cite{DBLP:journals/corr/abs-2007-06778}. Consider the first example, the model places the \<Horsepower> field in the \<SELECT> clause, while the question asks for ``the model of the car''. In the second example, the model predicts \<SELECT SUM(Population), AVG(Population)> while the question asks for \emph{total population and average area of countries}. We think better modeling of compositionality in the natural language may reduce this type of errors. For example, modeling its span structure ~\cite{DBLP:journals/corr/abs-1907-10529,DBLP:journals/corr/abs-2009-06040} and constructing interpretable grounding with the DB schema.


\section{Conclusion}

We present \Ours, 
a powerful sequential architecture for modeling dependencies between natural language question and 
relational DBs in cross-DB semantic parsing. 
\Ours serializes the question and DB schema into a tagged sequence and maximally utilizes pre-trained LMs such as BERT to capture the linking between text mentions and the DB schema components. It uses anchor texts to further improve the alignment between the two cross-modal inputs. 
Combined with a simple sequential pointer-generator decoder with schema-consistency driven search space pruning, \Ours attained state-of-the-art performance on the widely used Spider and WikiSQL text-to-SQL benchmarks. 

Our analysis shows that \Ours is effective at generalizing over natural language variations and memorizing structural patterns. It achieves the upperbound score on WikiSQL and significantly outperforms previous work in the easy category of Spider. 
However, it struggles in compositional generalization and sometimes makes unexplainable mistakes. 
This indicates that when data is ample and the target logic form is shallow, sequence-to-sequence models are good choices for cross-DB semantic parsing, especially given the implementation is easier and decoding is efficient. For solving the general text-to-SQL problem and moving towards production, we plan to further improve compositional generalization and interpretability of the model. We also plan to study the application of \Ours and its extensions to other tasks that requires joint textual and tabular understanding such as weakly supervised semantic parsing and fact checking.

\section*{Acknowledgements}
We thank Yingbo Zhou for helpful discussions. We thank the anonymous reviewers and members of Salesforce Research for their thoughtful feedback. A significant part of the experiments were completed during the California Bay Area shelter-in-place order for COVID-19. Our heartful thanks go to all who worked hard to keep others safe and enjoy a well-functioning life during this challenging time. \includegraphics[width=0.04\textwidth]{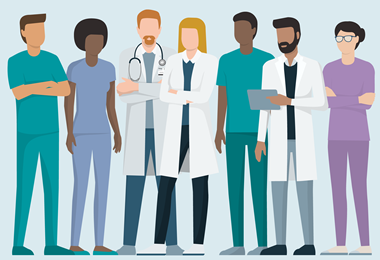}

\bibliography{emnlp2020}

\begin{thebibliography}{58}
\expandafter\ifx\csname natexlab\endcsname\relax\def\natexlab#1{#1}\fi

\bibitem[{Androutsopoulos et~al.(1995)Androutsopoulos, Ritchie, and
  Thanisch}]{androutsopoulos_ritchie_thanisch_1995}
I.~Androutsopoulos, G.D. Ritchie, and P.~Thanisch. 1995.
\newblock \href {https://doi.org/10.1017/S135132490000005X} {Natural language
  interfaces to databases – an introduction}.
\newblock \emph{Natural Language Engineering}, 1(1):29–81.

\bibitem[{Beltagy et~al.(2020)Beltagy, Peters, and
  Cohan}]{DBLP:journals/corr/abs-2004-05150}
Iz~Beltagy, Matthew~E. Peters, and Arman Cohan. 2020.
\newblock \href {http://arxiv.org/abs/2004.05150} {Longformer: The
  long-document transformer}.
\newblock \emph{CoRR}, abs/2004.05150.

\bibitem[{Bogin et~al.(2019{\natexlab{a}})Bogin, Berant, and
  Gardner}]{DBLP:conf/acl/BoginBG19}
Ben Bogin, Jonathan Berant, and Matt Gardner. 2019{\natexlab{a}}.
\newblock \href {https://doi.org/10.18653/v1/p19-1448} {Representing schema
  structure with graph neural networks for text-to-sql parsing}.
\newblock In \emph{Proceedings of the 57th Conference of the Association for
  Computational Linguistics, {ACL} 2019, Florence, Italy, July 28- August 2,
  2019, Volume 1: Long Papers}, pages 4560--4565. Association for Computational
  Linguistics.

\bibitem[{Bogin et~al.(2019{\natexlab{b}})Bogin, Gardner, and
  Berant}]{DBLP:conf/emnlp/BoginGB19}
Ben Bogin, Matt Gardner, and Jonathan Berant. 2019{\natexlab{b}}.
\newblock \href {https://doi.org/10.18653/v1/D19-1378} {Global reasoning over
  database structures for text-to-sql parsing}.
\newblock In \emph{Proceedings of the 2019 Conference on Empirical Methods in
  Natural Language Processing and the 9th International Joint Conference on
  Natural Language Processing, {EMNLP-IJCNLP} 2019, Hong Kong, China, November
  3-7, 2019}, pages 3657--3662. Association for Computational Linguistics.

\bibitem[{Choi et~al.(2020)Choi, Shin, Kim, and
  Shin}]{DBLP:journals/corr/abs-2004-03125}
DongHyun Choi, Myeong~Cheol Shin, EungGyun Kim, and Dong~Ryeol Shin. 2020.
\newblock {RYANSQL:} recursively applying sketch-based slot fillings for
  complex text-to-sql in cross-domain databases.
\newblock \emph{CoRR}, abs/2004.03125.

\bibitem[{Dahl et~al.(1994)Dahl, Bates, Brown, Fisher, Hunicke{-}Smith,
  Pallett, Pao, Rudnicky, and Shriberg}]{DBLP:conf/naacl/DahlBBFHPPRS94}
Deborah~A. Dahl, Madeleine Bates, Michael Brown, William~M. Fisher, Kate
  Hunicke{-}Smith, David~S. Pallett, Christine Pao, Alexander~I. Rudnicky, and
  Elizabeth Shriberg. 1994.
\newblock Expanding the scope of the {ATIS} task: The {ATIS-3} corpus.
\newblock In \emph{Human Language Technology, Proceedings of a Workshop held at
  Plainsboro, New Jerey, USA, March 8-11, 1994}.

\bibitem[{Devlin et~al.(2019)Devlin, Chang, Lee, and
  Toutanova}]{DBLP:conf/naacl/DevlinCLT19}
Jacob Devlin, Ming{-}Wei Chang, Kenton Lee, and Kristina Toutanova. 2019.
\newblock {BERT:} pre-training of deep bidirectional transformers for language
  understanding.
\newblock In \emph{Proceedings of the 2019 Conference of the North American
  Chapter of the Association for Computational Linguistics: Human Language
  Technologies, {NAACL-HLT} 2019, Minneapolis, MN, USA, June 2-7, 2019, Volume
  1}, pages 4171--4186.

\bibitem[{Dong and Lapata(2016)}]{DBLP:conf/acl/DongL16}
Li~Dong and Mirella Lapata. 2016.
\newblock \href {https://www.aclweb.org/anthology/P16-1004/} {Language to
  logical form with neural attention}.
\newblock In \emph{Proceedings of the 54th Annual Meeting of the Association
  for Computational Linguistics, {ACL} 2016, August 7-12, 2016, Berlin,
  Germany, Volume 1: Long Papers}. The Association for Computer Linguistics.

\bibitem[{Furrer et~al.(2020)Furrer, van Zee, Scales, and
  Sch{\"{a}}rli}]{DBLP:journals/corr/abs-2007-08970}
Daniel Furrer, Marc van Zee, Nathan Scales, and Nathanael Sch{\"{a}}rli. 2020.
\newblock \href {http://arxiv.org/abs/2007.08970} {Compositional generalization
  in semantic parsing: Pre-training vs. specialized architectures}.
\newblock \emph{CoRR}, abs/2007.08970.

\bibitem[{Gu et~al.(2016)Gu, Lu, Li, and Li}]{DBLP:conf/acl/GuLLL16}
Jiatao Gu, Zhengdong Lu, Hang Li, and Victor O.~K. Li. 2016.
\newblock \href {https://doi.org/10.18653/v1/p16-1154} {Incorporating copying
  mechanism in sequence-to-sequence learning}.
\newblock In \emph{Proceedings of the 54th Annual Meeting of the Association
  for Computational Linguistics, {ACL} 2016, August 7-12, 2016, Berlin,
  Germany, Volume 1: Long Papers}. The Association for Computer Linguistics.

\bibitem[{Guo et~al.(2019)Guo, Zhan, Gao, Xiao, Lou, Liu, and
  Zhang}]{DBLP:conf/acl/GuoZGXLLZ19}
Jiaqi Guo, Zecheng Zhan, Yan Gao, Yan Xiao, Jian{-}Guang Lou, Ting Liu, and
  Dongmei Zhang. 2019.
\newblock Towards complex text-to-sql in cross-domain database with
  intermediate representation.
\newblock In \emph{Proceedings of the 57th Conference of the Association for
  Computational Linguistics, {ACL} 2019, Florence, Italy, July 28- August 2,
  2019, Volume 1: Long Papers}, pages 4524--4535.

\bibitem[{Guo and Gao(2019)}]{DBLP:journals/corr/abs-1910-07179}
Tong Guo and Huilin Gao. 2019.
\newblock \href {http://arxiv.org/abs/1910.07179} {Content enhanced bert-based
  text-to-sql generation}.
\newblock \emph{CoRR}, abs/1910.07179.

\bibitem[{He et~al.(2019{\natexlab{a}})He, Mao, Chakrabarti, and
  Chen}]{DBLP:journals/corr/abs-1908-08113}
Pengcheng He, Yi~Mao, Kaushik Chakrabarti, and Weizhu Chen. 2019{\natexlab{a}}.
\newblock \href {http://arxiv.org/abs/1908.08113} {{X-SQL:} reinforce schema
  representation with context}.
\newblock \emph{CoRR}, abs/1908.08113.

\bibitem[{He et~al.(2019{\natexlab{b}})He, Mao, Chakrabarti, and
  Chen}]{he2019x}
Pengcheng He, Yi~Mao, Kaushik Chakrabarti, and Weizhu Chen. 2019{\natexlab{b}}.
\newblock X-sql: reinforce schema representation with context.
\newblock \emph{arXiv preprint arXiv:1908.08113}.

\bibitem[{Hemphill et~al.(1990)Hemphill, Godfrey, and
  Doddington}]{DBLP:conf/naacl/HemphillGD90}
Charles~T. Hemphill, John~J. Godfrey, and George~R. Doddington. 1990.
\newblock The {ATIS} spoken language systems pilot corpus.
\newblock In \emph{Speech and Natural Language: Proceedings of a Workshop Held
  at Hidden Valley, Pennsylvania, USA, June 24-27, 1990}.

\bibitem[{Herzig and Berant(2020)}]{DBLP:journals/corr/abs-2009-06040}
Jonathan Herzig and Jonathan Berant. 2020.
\newblock \href {http://arxiv.org/abs/2009.06040} {Span-based semantic parsing
  for compositional generalization}.
\newblock \emph{CoRR}, abs/2009.06040.

\bibitem[{Herzig et~al.(2020)Herzig, Nowak, Müller, Piccinno, and
  Eisenschlos}]{tapas}
Jonathan Herzig, Paweł~Krzysztof Nowak, Thomas Müller, Francesco Piccinno,
  and Julian~Martin Eisenschlos. 2020.
\newblock \href {https://arxiv.org/abs/2004.02349} {Tapas: Weakly supervised
  table parsing via pre-training}.
\newblock In \emph{Proceedings of the 58th Annual Meeting of the Association
  for Computational Linguistics (Volume 1: Long Papers)}, Seattle, Washington,
  United States.
\newblock To appear.

\bibitem[{Hochreiter and Schmidhuber(1997)}]{DBLP:journals/neco/HochreiterS97}
Sepp Hochreiter and J{\"{u}}rgen Schmidhuber. 1997.
\newblock \href {https://doi.org/10.1162/neco.1997.9.8.1735} {Long short-term
  memory}.
\newblock \emph{Neural Computation}, 9(8):1735--1780.

\bibitem[{Hwang et~al.(2019)Hwang, Yim, Park, and
  Seo}]{DBLP:journals/corr/abs-1902-01069}
Wonseok Hwang, Jinyeung Yim, Seunghyun Park, and Minjoon Seo. 2019.
\newblock \href {http://arxiv.org/abs/1902.01069} {A comprehensive exploration
  on wikisql with table-aware word contextualization}.
\newblock \emph{CoRR}, abs/1902.01069.

\bibitem[{Joshi et~al.(2019)Joshi, Chen, Liu, Weld, Zettlemoyer, and
  Levy}]{DBLP:journals/corr/abs-1907-10529}
Mandar Joshi, Danqi Chen, Yinhan Liu, Daniel~S. Weld, Luke Zettlemoyer, and
  Omer Levy. 2019.
\newblock \href {http://arxiv.org/abs/1907.10529} {Spanbert: Improving
  pre-training by representing and predicting spans}.
\newblock \emph{CoRR}, abs/1907.10529.

\bibitem[{Kelkar et~al.(2020)Kelkar, Relan, Bhardwaj, Vaichal, and
  Relan}]{kelkar2020bertrand}
Amol Kelkar, Rohan Relan, Vaishali Bhardwaj, Saurabh Vaichal, and Peter Relan.
  2020.
\newblock Bertrand-dr: Improving text-to-sql using a discriminative re-ranker.
\newblock \emph{arXiv preprint arXiv:2002.00557}.

\bibitem[{Kingma and Ba(2015)}]{DBLP:journals/corr/KingmaB14}
Diederik~P. Kingma and Jimmy Ba. 2015.
\newblock \href {http://arxiv.org/abs/1412.6980} {Adam: {A} method for
  stochastic optimization}.
\newblock In \emph{3rd International Conference on Learning Representations,
  {ICLR} 2015, San Diego, CA, USA, May 7-9, 2015, Conference Track
  Proceedings}.

\bibitem[{Korhonen et~al.(2019)Korhonen, Traum, and
  M{\`{a}}rquez}]{DBLP:conf/acl/2019-1}
Anna Korhonen, David~R. Traum, and Llu{\'{\i}}s M{\`{a}}rquez, editors. 2019.
\newblock \href {https://www.aclweb.org/anthology/volumes/P19-1/}
  {\emph{Proceedings of the 57th Conference of the Association for
  Computational Linguistics, {ACL} 2019, Florence, Italy, July 28- August 2,
  2019, Volume 1: Long Papers}}. Association for Computational Linguistics.

\bibitem[{Liang et~al.(2018)Liang, Norouzi, Berant, Le, and
  Lao}]{DBLP:conf/nips/LiangNBLL18}
Chen Liang, Mohammad Norouzi, Jonathan Berant, Quoc~V. Le, and Ni~Lao. 2018.
\newblock \href
  {http://papers.nips.cc/paper/8204-memory-augmented-policy-optimization-for-program-synthesis-and-semantic-parsing}
  {Memory augmented policy optimization for program synthesis and semantic
  parsing}.
\newblock In \emph{Advances in Neural Information Processing Systems 31: Annual
  Conference on Neural Information Processing Systems 2018, NeurIPS 2018, 3-8
  December 2018, Montr{\'{e}}al, Canada}, pages 10015--10027.

\bibitem[{Lin et~al.(2020)Lin, Socher, and Xiong}]{DBLP:conf/emnlp/LinSX20}
Xi~Victoria Lin, Richard Socher, and Caiming Xiong. 2020.
\newblock \href {https://www.aclweb.org/anthology/2020.findings-emnlp.438/}
  {Bridging textual and tabular data for cross-domain text-to-sql semantic
  parsing}.
\newblock In \emph{Proceedings of the 2020 Conference on Empirical Methods in
  Natural Language Processing: Findings, {EMNLP} 2020, Online Event, 16-20
  November 2020}, pages 4870--4888. Association for Computational Linguistics.

\bibitem[{Lin et~al.(2018)Lin, Wang, Zettlemoyer, and
  Ernst}]{DBLP:conf/lrec/LinWZE18}
Xi~Victoria Lin, Chenglong Wang, Luke Zettlemoyer, and Michael~D. Ernst. 2018.
\newblock \href
  {http://www.lrec-conf.org/proceedings/lrec2018/summaries/1021.html} {Nl2bash:
  {A} corpus and semantic parser for natural language interface to the linux
  operating system}.
\newblock In \emph{Proceedings of the Eleventh International Conference on
  Language Resources and Evaluation, {LREC} 2018, Miyazaki, Japan, May 7-12,
  2018}. European Language Resources Association {(ELRA)}.

\bibitem[{Liu et~al.(2019{\natexlab{a}})Liu, Luo, Yang, Wu, Chang, and
  Sui}]{DBLP:conf/acl/LiuLYWCS19}
Tianyu Liu, Fuli Luo, Pengcheng Yang, Wei Wu, Baobao Chang, and Zhifang Sui.
  2019{\natexlab{a}}.
\newblock \href {https://doi.org/10.18653/v1/p19-1600} {Towards comprehensive
  description generation from factual attribute-value tables}.
\newblock In  \cite{DBLP:conf/acl/2019-1}, pages 5985--5996.

\bibitem[{Liu et~al.(2019{\natexlab{b}})Liu, He, Chen, and
  Gao}]{liu-etal-2019-multi}
Xiaodong Liu, Pengcheng He, Weizhu Chen, and Jianfeng Gao. 2019{\natexlab{b}}.
\newblock \href {https://doi.org/10.18653/v1/P19-1441} {Multi-task deep neural
  networks for natural language understanding}.
\newblock In \emph{Proceedings of the 57th Annual Meeting of the Association
  for Computational Linguistics}, pages 4487--4496, Florence, Italy.
  Association for Computational Linguistics.

\bibitem[{Liu et~al.(2019{\natexlab{c}})Liu, Ott, Goyal, Du, Joshi, Chen, Levy,
  Lewis, Zettlemoyer, and Stoyanov}]{DBLP:journals/corr/abs-1907-11692}
Yinhan Liu, Myle Ott, Naman Goyal, Jingfei Du, Mandar Joshi, Danqi Chen, Omer
  Levy, Mike Lewis, Luke Zettlemoyer, and Veselin Stoyanov. 2019{\natexlab{c}}.
\newblock \href {http://arxiv.org/abs/1907.11692} {Roberta: {A} robustly
  optimized {BERT} pretraining approach}.
\newblock \emph{CoRR}, abs/1907.11692.

\bibitem[{Lyu et~al.(2020)Lyu, Chakrabarti, Hathi, Kundu, Zhang, and
  Chen}]{lyu2020hybrid}
Qin Lyu, Kaushik Chakrabarti, Shobhit Hathi, Souvik Kundu, Jianwen Zhang, and
  Zheng Chen. 2020.
\newblock \href
  {https://www.microsoft.com/en-us/research/publication/hybrid-ranking-network-for-text-to-sql/}
  {Hybrid ranking network for text-to-sql}.
\newblock Technical Report MSR-TR-2020-7, Microsoft Dynamics 365 AI.

\bibitem[{Ma et~al.(2020)Ma, Yan, Pang, Zhang, and Shen}]{ma-etal-2020-mention}
Jianqiang Ma, Zeyu Yan, Shuai Pang, Yang Zhang, and Jianping Shen. 2020.
\newblock \href {https://doi.org/10.18653/v1/2020.emnlp-main.563} {Mention
  extraction and linking for {SQL} query generation}.
\newblock In \emph{Proceedings of the 2020 Conference on Empirical Methods in
  Natural Language Processing (EMNLP)}, pages 6936--6942, Online. Association
  for Computational Linguistics.

\bibitem[{Raffel et~al.(2020)Raffel, Shazeer, Roberts, Lee, Narang, Matena,
  Zhou, Li, and Liu}]{DBLP:journals/jmlr/RaffelSRLNMZLL20}
Colin Raffel, Noam Shazeer, Adam Roberts, Katherine Lee, Sharan Narang, Michael
  Matena, Yanqi Zhou, Wei Li, and Peter~J. Liu. 2020.
\newblock \href {http://jmlr.org/papers/v21/20-074.html} {Exploring the limits
  of transfer learning with a unified text-to-text transformer}.
\newblock \emph{J. Mach. Learn. Res.}, 21:140:1--140:67.

\bibitem[{Rob and Coronel(1995)}]{DBLP:books/daglib/0078665}
Peter Rob and Carlos Coronel. 1995.
\newblock \emph{Database systems - design, implementation, and management {(2.}
  ed.)}.
\newblock Boyd and Fraser.

\bibitem[{Rubin and Berant(2020)}]{DBLP:journals/corr/abs-2010-12412}
Ohad Rubin and Jonathan Berant. 2020.
\newblock \href {http://arxiv.org/abs/2010.12412} {Smbop: Semi-autoregressive
  bottom-up semantic parsing}.
\newblock \emph{CoRR}, abs/2010.12412.

\bibitem[{See et~al.(2017)See, Liu, and Manning}]{DBLP:conf/acl/SeeLM17}
Abigail See, Peter~J. Liu, and Christopher~D. Manning. 2017.
\newblock Get to the point: Summarization with pointer-generator networks.
\newblock In \emph{Proceedings of the 55th Annual Meeting of the Association
  for Computational Linguistics, {ACL} 2017, Vancouver, Canada, July 30 -
  August 4, Volume 1: Long Papers}, pages 1073--1083.

\bibitem[{Shaw et~al.(2020)Shaw, Chang, Pasupat, and
  Toutanova}]{DBLP:journals/corr/abs-2010-12725}
Peter Shaw, Ming{-}Wei Chang, Panupong Pasupat, and Kristina Toutanova. 2020.
\newblock \href {http://arxiv.org/abs/2010.12725} {Compositional generalization
  and natural language variation: Can a semantic parsing approach handle both?}
\newblock \emph{CoRR}, abs/2010.12725.

\bibitem[{Shaw et~al.(2019)Shaw, Massey, Chen, Piccinno, and
  Altun}]{DBLP:conf/acl/ShawMCPA19}
Peter Shaw, Philip Massey, Angelica Chen, Francesco Piccinno, and Yasemin
  Altun. 2019.
\newblock \href {https://doi.org/10.18653/v1/p19-1010} {Generating logical
  forms from graph representations of text and entities}.
\newblock In \emph{Proceedings of the 57th Conference of the Association for
  Computational Linguistics, {ACL} 2019, Florence, Italy, July 28- August 2,
  2019, Volume 1: Long Papers}, pages 95--106. Association for Computational
  Linguistics.

\bibitem[{Shaw et~al.(2018)Shaw, Uszkoreit, and
  Vaswani}]{DBLP:conf/naacl/ShawUV18}
Peter Shaw, Jakob Uszkoreit, and Ashish Vaswani. 2018.
\newblock \href {https://doi.org/10.18653/v1/n18-2074} {Self-attention with
  relative position representations}.
\newblock In \emph{Proceedings of the 2018 Conference of the North American
  Chapter of the Association for Computational Linguistics: Human Language
  Technologies, NAACL-HLT, New Orleans, Louisiana, USA, June 1-6, 2018, Volume
  2 (Short Papers)}, pages 464--468. Association for Computational Linguistics.

\bibitem[{Strubell et~al.(2018)Strubell, Verga, Andor, Weiss, and
  McCallum}]{DBLP:conf/emnlp/StrubellVAWM18}
Emma Strubell, Patrick Verga, Daniel Andor, David Weiss, and Andrew McCallum.
  2018.
\newblock \href {https://www.aclweb.org/anthology/D18-1548/}
  {Linguistically-informed self-attention for semantic role labeling}.
\newblock In \emph{Proceedings of the 2018 Conference on Empirical Methods in
  Natural Language Processing, Brussels, Belgium, October 31 - November 4,
  2018}, pages 5027--5038. Association for Computational Linguistics.

\bibitem[{Suhr et~al.(2020)Suhr, Chang, Shaw, and Lee}]{Suhr2020}
Alane Suhr, Ming-Wei Chang, Peter Shaw, and Kenton Lee. 2020.
\newblock Exploring unexplored generalization challenges for cross-database
  semantic parsing.
\newblock In \emph{The 58th annual meeting of the Association for Computational
  Linguistics (ACL)}.

\bibitem[{Tu et~al.(2020)Tu, Lalwani, Gella, and
  He}]{DBLP:journals/corr/abs-2007-06778}
Lifu Tu, Garima Lalwani, Spandana Gella, and He~He. 2020.
\newblock \href {http://arxiv.org/abs/2007.06778} {An empirical study on
  robustness to spurious correlations using pre-trained language models}.
\newblock \emph{CoRR}, abs/2007.06778.

\bibitem[{Vaswani et~al.(2017)Vaswani, Shazeer, Parmar, Uszkoreit, Jones,
  Gomez, Kaiser, and Polosukhin}]{DBLP:conf/nips/VaswaniSPUJGKP17}
Ashish Vaswani, Noam Shazeer, Niki Parmar, Jakob Uszkoreit, Llion Jones,
  Aidan~N. Gomez, Lukasz Kaiser, and Illia Polosukhin. 2017.
\newblock \href {http://papers.nips.cc/paper/7181-attention-is-all-you-need}
  {Attention is all you need}.
\newblock In \emph{Advances in Neural Information Processing Systems 30: Annual
  Conference on Neural Information Processing Systems 2017, 4-9 December 2017,
  Long Beach, CA, {USA}}, pages 5998--6008.

\bibitem[{Vig(2019)}]{vig2019transformervis}
Jesse Vig. 2019.
\newblock \href {https://arxiv.org/abs/1906.05714} {A multiscale visualization
  of attention in the transformer model}.
\newblock \emph{arXiv preprint arXiv:1906.05714}.

\bibitem[{Wang et~al.(2019)Wang, Shin, Liu, Polozov, and
  Richardson}]{Wang2019RATSQLRS}
Bailin Wang, Richard Shin, Xiaodong Liu, Oleksandr Polozov, and Margot
  Richardson. 2019.
\newblock Rat-sql: Relation-aware schema encoding and linking for text-to-sql
  parsers.
\newblock \emph{ArXiv}, abs/1911.04942.

\bibitem[{Wang et~al.(2018)Wang, Huang, Polozov, Brockschmidt, and
  Singh}]{DBLP:journals/corr/abs-1807-03100}
Chenglong Wang, Po{-}Sen Huang, Alex Polozov, Marc Brockschmidt, and Rishabh
  Singh. 2018.
\newblock \href {http://arxiv.org/abs/1807.03100} {Execution-guided neural
  program decoding}.
\newblock \emph{CoRR}, abs/1807.03100.

\bibitem[{Wolf et~al.(2019)Wolf, Debut, Sanh, Chaumond, Delangue, Moi, Cistac,
  Rault, Louf, Funtowicz, and Brew}]{Wolf2019HuggingFacesTS}
Thomas Wolf, Lysandre Debut, Victor Sanh, Julien Chaumond, Clement Delangue,
  Anthony Moi, Pierric Cistac, Tim Rault, R'emi Louf, Morgan Funtowicz, and
  Jamie Brew. 2019.
\newblock Huggingface's transformers: State-of-the-art natural language
  processing.
\newblock \emph{ArXiv}, abs/1910.03771.

\bibitem[{Yao et~al.(2020)Yao, Tang, Yih, Sun, and
  Su}]{DBLP:conf/emnlp/YaoTYSS20}
Ziyu Yao, Yiqi Tang, Wen{-}tau Yih, Huan Sun, and Yu~Su. 2020.
\newblock \href {https://www.aclweb.org/anthology/2020.emnlp-main.559/} {An
  imitation game for learning semantic parsers from user interaction}.
\newblock In \emph{Proceedings of the 2020 Conference on Empirical Methods in
  Natural Language Processing, {EMNLP} 2020, Online, November 16-20, 2020},
  pages 6883--6902. Association for Computational Linguistics.

\bibitem[{Yavuz et~al.(2018)Yavuz, Gur, Su, and
  Yan}]{DBLP:conf/emnlp/YavuzG0Y18}
Semih Yavuz, Izzeddin Gur, Yu~Su, and Xifeng Yan. 2018.
\newblock \href {https://doi.org/10.18653/v1/d18-1197} {What it takes to
  achieve 100 percent condition accuracy on wikisql}.
\newblock In \emph{Proceedings of the 2018 Conference on Empirical Methods in
  Natural Language Processing, Brussels, Belgium, October 31 - November 4,
  2018}, pages 1702--1711. Association for Computational Linguistics.

\bibitem[{Yin and Neubig(2017)}]{DBLP:conf/acl/YinN17}
Pengcheng Yin and Graham Neubig. 2017.
\newblock A syntactic neural model for general-purpose code generation.
\newblock In \emph{Proceedings of the 55th Annual Meeting of the Association
  for Computational Linguistics, {ACL} 2017, Vancouver, Canada, July 30 -
  August 4, Volume 1: Long Papers}, pages 440--450.

\bibitem[{Yin et~al.(2020)Yin, Neubig, Yih, and
  Riedel}]{DBLP:journals/corr/abs-2005-08314}
Pengcheng Yin, Graham Neubig, Wen{-}tau Yih, and Sebastian Riedel. 2020.
\newblock \href {http://arxiv.org/abs/2005.08314} {Tabert: Pretraining for
  joint understanding of textual and tabular data}.
\newblock \emph{CoRR}, abs/2005.08314.

\bibitem[{Yu et~al.(2018)Yu, Yasunaga, Yang, Zhang, Wang, Li, and
  Radev}]{DBLP:conf/emnlp/YuYYZWLR18}
Tao Yu, Michihiro Yasunaga, Kai Yang, Rui Zhang, Dongxu Wang, Zifan Li, and
  Dragomir~R. Radev. 2018.
\newblock Syntaxsqlnet: Syntax tree networks for complex and cross-domain
  text-to-sql task.
\newblock In \emph{Proceedings of the 2018 Conference on Empirical Methods in
  Natural Language Processing, Brussels, Belgium, October 31 - November 4,
  2018}, pages 1653--1663.

\bibitem[{Yu et~al.(2019{\natexlab{a}})Yu, Zhang, Er, Li, Xue, Pang, Lin, Tan,
  Shi, Li, Jiang, Yasunaga, Shim, Chen, Fabbri, Li, Chen, Zhang, Dixit, Zhang,
  Xiong, Socher, Lasecki, and Radev}]{DBLP:journals/corr/abs-1909-05378}
Tao Yu, Rui Zhang, Heyang Er, Suyi Li, Eric Xue, Bo~Pang, Xi~Victoria Lin,
  Yi~Chern Tan, Tianze Shi, Zihan Li, Youxuan Jiang, Michihiro Yasunaga,
  Sungrok Shim, Tao Chen, Alexander~Richard Fabbri, Zifan Li, Luyao Chen, Yuwen
  Zhang, Shreya Dixit, Vincent Zhang, Caiming Xiong, Richard Socher, Walter~S.
  Lasecki, and Dragomir~R. Radev. 2019{\natexlab{a}}.
\newblock \href {http://arxiv.org/abs/1909.05378} {Cosql: {A} conversational
  text-to-sql challenge towards cross-domain natural language interfaces to
  databases}.
\newblock \emph{CoRR}, abs/1909.05378.

\bibitem[{Yu et~al.(2019{\natexlab{b}})Yu, Zhang, Yasunaga, Tan, Lin, Li, Er,
  Li, Pang, Chen, Ji, Dixit, Proctor, Shim, Kraft, Zhang, Xiong, Socher, and
  Radev}]{DBLP:conf/acl/YuZYTLLELPCJDPS19}
Tao Yu, Rui Zhang, Michihiro Yasunaga, Yi~Chern Tan, Xi~Victoria Lin, Suyi Li,
  Heyang Er, Irene Li, Bo~Pang, Tao Chen, Emily Ji, Shreya Dixit, David
  Proctor, Sungrok Shim, Jonathan Kraft, Vincent Zhang, Caiming Xiong, Richard
  Socher, and Dragomir~R. Radev. 2019{\natexlab{b}}.
\newblock \href {https://www.aclweb.org/anthology/P19-1443/} {Sparc:
  Cross-domain semantic parsing in context}.
\newblock In  \cite{DBLP:conf/acl/2019-1}, pages 4511--4523.

\bibitem[{Zelle and Mooney(1996)}]{DBLP:conf/aaai/ZelleM96}
John~M. Zelle and Raymond~J. Mooney. 1996.
\newblock Learning to parse database queries using inductive logic programming.
\newblock In \emph{Proceedings of the Thirteenth National Conference on
  Artificial Intelligence and Eighth Innovative Applications of Artificial
  Intelligence Conference, {AAAI} 96, {IAAI} 96, Portland, Oregon, USA, August
  4-8, 1996, Volume 2.}, pages 1050--1055.

\bibitem[{Zeng et~al.(2020)Zeng, Lin, Hoi, Socher, Xiong, Lyu, and
  King}]{DBLP:conf/acl/ZengLHSXLK20}
Jichuan Zeng, Xi~Victoria Lin, Steven C.~H. Hoi, Richard Socher, Caiming Xiong,
  Michael~R. Lyu, and Irwin King. 2020.
\newblock \href {https://www.aclweb.org/anthology/2020.acl-demos.24/} {Photon:
  {A} robust cross-domain text-to-sql system}.
\newblock In \emph{Proceedings of the 58th Annual Meeting of the Association
  for Computational Linguistics: System Demonstrations, {ACL} 2020, Online,
  July 5-10, 2020}, pages 204--214. Association for Computational Linguistics.

\bibitem[{Zettlemoyer and Collins(2005)}]{DBLP:conf/uai/ZettlemoyerC05}
Luke~S. Zettlemoyer and Michael Collins. 2005.
\newblock \href
  {https://dslpitt.org/uai/displayArticleDetails.jsp?mmnu=1\&smnu=2\&article\_id=1209\&proceeding\_id=21}
  {Learning to map sentences to logical form: Structured classification with
  probabilistic categorial grammars}.
\newblock In \emph{{UAI} '05, Proceedings of the 21st Conference in Uncertainty
  in Artificial Intelligence, Edinburgh, Scotland, July 26-29, 2005}, pages
  658--666. {AUAI} Press.

\bibitem[{Zhang et~al.(2019)Zhang, Yu, Er, Shim, Xue, Lin, Shi, Xiong, Socher,
  and Radev}]{DBLP:conf/emnlp/YuYYZWLR19}
Rui Zhang, Tao Yu, Heyang Er, Sungrok Shim, Eric Xue, Xi~Victoria Lin, Tianze
  Shi, Caiming Xiong, Richard Socher, and Dragomir~R. Radev. 2019.
\newblock \href {http://arxiv.org/abs/1909.00786} {Editing-based {SQL} query
  generation for cross-domain context-dependent questions}.
\newblock \emph{CoRR}, abs/1909.00786.

\bibitem[{Zhong et~al.(2017)Zhong, Xiong, and
  Socher}]{DBLP:journals/corr/abs-1709-00103}
Victor Zhong, Caiming Xiong, and Richard Socher. 2017.
\newblock Seq2sql: Generating structured queries from natural language using
  reinforcement learning.
\newblock \emph{CoRR}, abs/1709.00103.

\end{thebibliography}
\bibliographystyle{acl_natbib}

\newpage
\appendix

\setcounter{table}{0}
\renewcommand{\thetable}{A\arabic{table}}
\setcounter{figure}{0}
\renewcommand{\thefigure}{A\arabic{figure}}

\section{Appendix}
\label{sec:appendix}

\subsection{Examples of SQL queries with clauses arranged in execution order}
We show more examples of complex SQL queries with their clauses arranged in written order vs. execution order in Table~\ref{tab:execution_order}.

\begin{table*}[t]
    \centering
    \scalebox{0.84}{
    \noindent\begin{tabularx}{\linewidth}{X}
         \toprule
         \underline{Written:} \<\textbf{SELECT} rid \textbf{FROM} routes \textbf{WHERE} dst\_apid \textbf{IN} (\textbf{SELECT} apid \textbf{FROM} airports \textbf{WHERE} country = 'United States') \textbf{AND} src\_apid \textbf{IN} (\textbf{SELECT} apid \textbf{FROM} airports \textbf{WHERE} country  =  'United States')> \\
         \underline{Exec:} \<\textbf{FROM} routes \textbf{WHERE} dst\_apid \textbf{IN} (\textbf{FROM} airports \textbf{WHERE} country = 'United States' \textbf{SELECT} apid) \textbf{AND} src\_apid \textbf{IN} (\textbf{FROM} airports \textbf{WHERE} country = 'United States' \textbf{SELECT} apid) \textbf{SELECT} rid> \\
         \\
         \underline{Written:} \<\textbf{SELECT} t3.name \textbf{FROM} publication\_keyword \textbf{AS} t4 \textbf{JOIN} keyword \textbf{AS} t1 \textbf{ON} t4.kid = t1.kid \textbf{JOIN} publication \textbf{AS} t2 \textbf{ON} t2.pid = t4.pid \textbf{JOIN} journal \textbf{AS} t3 \textbf{ON} t2.jid = t3.jid \textbf{WHERE} t1.keyword = "Relational Database" \textbf{GROUP BY} t3.name HAVING \textbf{COUNT}(\textbf{DISTINCT} t2.title) = 60> \\
         \underline{Exec:} \<\textbf{FROM} publication\_keyword \textbf{AS} t4 \textbf{JOIN} keyword \textbf{AS} t1 \textbf{ON} t4.kid = t1.kid \textbf{JOIN} publication \textbf{AS} t2 \textbf{ON} t2.pid = t4.pid \textbf{JOIN} journal \textbf{AS} t3 \textbf{ON} t2.jid = t3.jid \textbf{WHERE} t1.keyword = "Relational Database" \textbf{GROUP BY} t3.name HAVING \textbf{COUNT}(\textbf{DISTINCT} t2.title) = 60 \textbf{SELECT} t3.name> \\
         \\
         \underline{Written:} \<\textbf{SELECT} \textbf{COUNT}(\textbf{DISTINCT} state) \textbf{FROM} college \textbf{WHERE} enr  <  (\textbf{SELECT} \textbf{AVG}(enr) \textbf{FROM} college)> \\
         \underline{Exec:} \<\textbf{FROM} college \textbf{WHERE} enr < (\textbf{FROM} college \textbf{SELECT} \textbf{AVG}(enr)) \textbf{SELECT} \textbf{COUNT}(\textbf{DISTINCT} state)> \\
         \\
         \underline{Written:} \<\textbf{SELECT} \textbf{DISTINCT} T1.LName \textbf{FROM} STUDENT \textbf{AS} T1 \textbf{JOIN} VOTING\_RECORD \textbf{AS} T2 \textbf{ON} T1.StuID  =  PRESIDENT\_Vote \textbf{EXCEPT} \textbf{SELECT} \textbf{DISTINCT} LName \textbf{FROM} STUDENT \textbf{WHERE} Advisor  =  "2192"> \\
         \underline{Exec:} \<\textbf{FROM} STUDENT \textbf{AS} T1 \textbf{JOIN} VOTING\_RECORD \textbf{AS} T2 \textbf{ON} T1.StuID = PRESIDENT\_Vote \textbf{SELECT} \textbf{DISTINCT} T1.LName \textbf{EXCEPT} \textbf{FROM} STUDENT \textbf{WHERE} Advisor = 2192 \textbf{SELECT} \textbf{DISTINCT} LName> \\
         \bottomrule
    \end{tabularx}}
    \caption{Examples of complex SQL queries with clauses in the normal order and the DB execution order.}
    \label{tab:execution_order}
\end{table*}

\begin{table*}[t]
    \centering
    \scalebox{0.86}{
    \noindent\begin{tabularx}{\linewidth}{X}
        \toprule
           \<\textbf{FROM} STUDENT \textbf{JOIN} VOTING\_RECORD \textbf{ON} STUDENT.StuID = VOTING\_RECORD.PRESIDENT\_Vote \textbf{SELECT} \textbf{DISTINCT} STUDENT.LName \textbf{EXCEPT}\textbf{FROM} STUDENT \textbf{WHERE} STUDENT.Advisor = 2192 \textbf{SELECT} \textbf{DISTINCT} VOTING\_RECORD.PRESIDENT\_Vote> \\
        \bottomrule  
    \end{tabularx}}
    \caption{An example sequence satisfies the condition of Lemma~\ref{lemma:field_ordering} but violates schema consistency. Here the field \<VOTING\_RECORD.PRESIDENT\_Vote> in the second sub-query is out of scope.}
    \label{tab:lemma_1_exception}
\end{table*}

\subsection{Selective read decoder extension}
\label{sec:selective_read}
The selective read operation was introduced by~\citet{DBLP:conf/acl/GuLLL16}.
It extends the input state to the decoder LSTM with the corresponding encoder hidden states of the tokens being copied. This way the decoder was provided information on which part of the input has been copied.

Specically, we modified the input state\footnote{The original formulation by~\citet{DBLP:conf/acl/GuLLL16} does not contain the $(1-p^t_\text{gen})$ term in Equation~\ref{eq:selective-read}. We introduce this term as for some tokens there is ambiguity regarding whether the token is copied or generated from the decoder vocabulary.} of our decoder LSTM to the following:
\begin{equation}
    \label{eq:selective-read}    
    \vect{y}_t=[\vect{e}_{t-1};(1-p^t_\text{gen})\cdot\zeta_{t-1}]\in\Real^{2n},
\end{equation}
where $p^t_\text{gen}$ is the scalar probability that a token is copied at step $t$. $\vect{e}_{t-1}\in\Real^n$ is either the embedding of a generated vocabulary token or a learned vector indicating if a table, field or question token is copied in step $t-1$. $\zeta_{t-1}\in\Real^n$ is the \emph{selective read} vector, which is a weighted sum of the encoder hidden states corresponding to the tokens copied in step $t-1$:
\begin{equation}
\small
\begin{split}
	\zeta(y_{t-1}) = \sum_{j=1}^{|\question|+|\schema|}\rho_{t-1, j} \vect{h}_j
	;\quad
	\rho_{t-1, j} = \left \{ \begin{matrix} 
	     \dfrac{1}{K} \alpha_{t-1,j}^{(H)},
	     \quad & \Tilde{X}_j = y_{t-1}\\
		0 \quad & \text{otherwise}
	\end{matrix}\right. \\
\end{split} \vspace{-7pt}
\end{equation}
 Here $K = \sum_{j:\Tilde{X}_j=y_{t-1}} \alpha_{t-1,j}^{(H)}$ is a normalization term considering there may be multiple positions equals to $y_{t-1}$ in $\tilde{X}$.
 
\subsection{Anchor text selection}
\label{sec:picklist-eval}

We convert the question and field values into lower cased character sequences and compute the longest sub-sequence match with heuristically determined matching boundaries. For example, the sentence ``how many students keep cats as pets?'' matches with the cell value ``cat'' ($s_c$) and the matched sub-string is ``cat'' ($s_m$). We further search the question starting from the start and end character indices $i, j$ of $s_m$ in the question to make sure that word boundaries can be detected within $i-2$ to $j+2$, otherwise the match is invalidated. This excludes matches which are sub-strings of the question words, e.g. ``cat'' vs. ``category''. Denoting matched whole-word phrase in the question as $s_q$, we define the question match score and cell match score as
\begin{align}
    \beta_q &= |s_m|/|s_q| \\
    \beta_c &= |s_c|/|s_q|
\end{align}

We define a coarse accuracy measurement to tune the question match score threshold $\theta_q$ and the cell match threshold $\theta_c$. Namely, given the list of matched anchor texts $\sPicklist$ obtained using the aforementioned procedure and the list of textual values $\mathcal{G}$ extracted from the ground truth SQL query, when compute the percentage of anchor texts appeared in $\mathcal{G}$ and the percentage of values in $\mathcal{G}$ that appeared in $\sPicklist$ as approximated precision ($p'$) and recall ($r'$). Note that this metrics does not evaluate if the matched anchor texts are associated with the correct field.

For $k=2$, we set $\theta_q=0.5$ and $\theta_c=0.8$. On the training set, the resulting $p'=73.7, r'=74.9$. 25.7\% examples have at least one anchor text match with 1.89 average number of matches per example among them. On the dev set, the resulting $p'=90.0, r'=92.2$. 30.9\% examples have at least one match with 1.73 average number of matches per example among them. The training set metrics are lower as some training databases do not have DB content files.

To quantify the effect of anchor text matching accuracy to the end-to-end performance, we run a set of experiments comparing \Ours performance w.r.t. different anchor text matching F1s. 
Our preliminary results show that with the same anchor text matching recall, varying the precision does not significantly change the end-to-end model performance. 

\hide{
\begin{table}[t]
    \centering
    \scalebox{0.78}{
    \begin{tabular}{llll}
        \toprule
        \bfseries Train & \bfseries Dev & \multicolumn{2}{c}{\bfseries Exact Set Match (\%)} \\
        \bfseries P/R/F1 & \bfseries P/R/F1 & Mean & Max \\
        \midrule
        74.9 / 80.9 / 75.7 & 89.8 / 93.7 / 90.2 & 68.2 \textpm~1.0 & 69.1 \\
        73.8 / 80.8 / 74.6 & 84.9 / 93.7 / 85.7 & \textbf{68.7 \textpm~0.2} & \textbf{69.2} \\
        72.2 / 80.9 / 73.2 & 77.5 / 93.7 / 78.8 & 67.9 \textpm~0.3 & 68.7 \\
        \bottomrule
    \end{tabular} }
    \caption{End-to-end performance of \Ours$_\text{L}$ w.r.t. anchor text matching F1s. We report the E-SM of each model averaged over 3 runs.}
    \label{tab:performance-anchor-text}
\end{table}
}

\hide{
\subsection{Performance by anchor text matching accuracy} 
We vary the thresholds $\theta_q$ and $\theta_c$ described in \S\ref{sec:picklist-eval} to create anchor text matching with different F1s. Table~\ref{tab:performance-anchor-text} shows the end-to-end performance of \Ours$_\text{L}$ w.r.t. different anchor text matching F1s.
}

\subsection{Performance by number of attention heads}
\label{sec:performance-attn-heads}
\begin{table}
    \centering
    \scalebox{0.9}{
    \begin{tabular}{lll}
        \toprule
        \bfseries \# & \multicolumn{2}{c}{\bfseries Exact Set Match (\%)} \\
        \bfseries attn heads & Mean & Max \\
        \midrule
        1-last & 67.4\textpm~0.3 & 67.7 \\
        2-last & \textbf{68.1}\textpm~0.8 & \textbf{69.1} \\
        4-last & \textbf{68.1}\textpm~0.5 & 68.8 \\
        8-last & 67.7\textpm~0.7 & 68.7 \\
        \hdashline
        2-mean & 67.8\textpm~0.6 & 68.8 \\
        \bottomrule
    \end{tabular} }
    \caption{End-to-end performance of \Ours$_\text{L}$ w.r.t. different number of attention heads between encoder and decoder. ``-last'' indicates the last attention head is used as the copy probability. ``-mean'' indicates the mean of all attention heads is used. We report the E-SM of each model averaged over 5 runs.}
    \label{tab:performance-attn-heads}
\end{table}

While multi-head attention between encoder and decoder is typically used in transformers~\cite{DBLP:conf/nips/VaswaniSPUJGKP17}, our experiments show they are effective for the \Ours model as well. Table~\ref{tab:performance-attn-heads} shows the performance of \Ours$_\text{L}$ w.r.t. different number of attention heads, where the attention probability computed by the last head is used as the copy probability. We saw that using more than 1 heads in general significantly improves over using only 1 head, where both the 2-head and 4-head attentions give best performance.

\subsection{The Linear-inverse-square-root (L-inv) learning rate decay function}
\label{sec:linv}
The linear ($\gamma_0-\alpha n$) and inverse-square-root ($\frac{\gamma_0}{\sqrt{n}}$) learning rate schedulers are commonly used for learning rate decay in neural network training\footnote{\url{https://fairseq.readthedocs.io/en/latest/lr_scheduler.html}}. 
The linear one decays slower in the beginning but slower in the end. The inverse-square-root one decays faster in the beginning but approaches 0 when $n\rightarrow\text{inf}$. We hence combine the two functions and propose a new learning rate scheduler that both decays fast in the beginning and also reaches 0 with finite $n$. The L-inv learning rate scheduler is defined as:
$$\frac{\gamma_0}{\sqrt{n}}-\beta n,$$
where $\beta=\frac{\gamma_0}{\sqrt{n_{\text{max}}}}$ and $n_{\text{max}}$ is the total number of back-propagation steps.

\subsection{Ensemble Modeling}
\label{sec:ensemble}

\begin{figure}[t]
    \centering
    \includegraphics[width=0.38\textwidth]{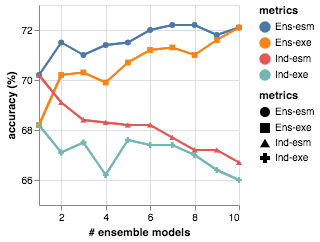}
    \caption{Performance ensemble models w.r.t. different \# models in the ensemble.}
    \label{fig:ensemble}
\end{figure}

As shown in \S\ref{sec:ablation}, the performance of \Ours on Spider is sensitive to the random seed. We train 10 different \Ours models with only differences in the random seeds. Figure~\ref{fig:ensemble} shows the performance of each individual model (sorted in decreasing exact set match), and the top-$k$ models ensembled using average step probabilities. 

The individual model performance variation is indeed large. The best and the worst models differ by 3.4 absolute points in E-SM, and 2.2 absolute points in execution accuracy.\footnote{In general the execution accuracy of our model is lower than the E-SM. We believe the execution accuracy can be further improved by copying the anchor texts during SQL generation.} We hypothesize that this is a result of both intrinsic model variance as well as error in the evaluation metrics. Considering the false negatives, the true model performance could have less variance. Combining models in general leads to better performance. In particular, combining the best model with the second best model improves the E-SM by 1.3 absolute points. Further combining with the weaker models still shows improvements, but the return is diminishing. The top-7 model ensemble achieves the best E-SM (72.2\%) and the top-10 model ensemble achieves the best execution accuracy (72.1\%).

\begin{table}[t]
    \centering
    \scalebox{0.9}{
    \begin{tabular}{c|c|c}
         & Best \cmark & Best \xmark \\
        \hline
        Worst \cmark & 61.2\% & 5.5\% \\
        Worst \xmark & 8.9\% & 24.4\% \\
    \end{tabular}
    }
    \caption{Comparison of the best and worst model obtained via different random seeds in terms of error overlap on the Spider dev set.}
    \label{tab:best_worst_overlap}
\end{table}
Table~\ref{tab:best_worst_overlap} shows the comparison between the best (70.2\%) and worst (66.7\%) models on the Spider dev set in terms of error overlap. For 61\% of dev set, both models predicted the corrected answer and for 24.4\% of dev set both models made a mistake. For 8.9\% of the examples, only the best model is correct, while for 5.5\% of the examples the worst model is correct. Manual examination shows that most of the examples where the two models evaluate differently are indeed different semantically.

\subsection{Performance by Database}
\label{sec:performance_by_db}
\begin{figure}[t]
    \centering
    \includegraphics[width=0.45\textwidth]{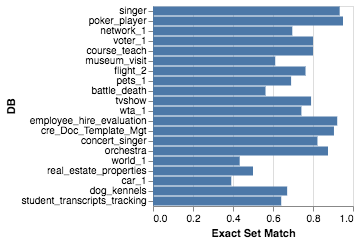}
    \caption{E-SM accuracy of \Ours by DB in Spider dev set. From top to bottom, the DBs are sorted by \# tables in the schema in ascending order.}
    \label{fig:esm_by_db}
\end{figure}

We further compute the E-SM accuracy of \Ours over different DBs in the Spider dev set. Figure~\ref{fig:esm_by_db} shows drastic performance differences across DBs. While \Ours achieves near perfect score on some, the performance is only 30\%-40\% on others. Performance does not always negatively correlates with the schema size. We hypothesize that the model scores better on DB schema similar to those seen during training and better characterization of the ``similarity'' between DB schema could help transfer learning.

\subsection{Visualizing fine-turned BERT attention of \Ours}
\label{sec:bert_attn_vis}

We visualize attention in the fine-tuned BERT layers of \Ours (with BERT-base-uncased) to qualitatively evaluate if the model functions as an effective text-DB encoder as we expect. We use the BERTViz library\footnote{\url{https://github.com/jessevig/bertviz}} developed by~\citet{vig2019transformervis}.

We perform the analysis on the smallest DB in the Spider dev set to ensure the attention graphs are readable. The DB consists of two tables, \<Poker\_ Player> and \<People> that store information of poker players and their match results. While the BERT attention is a 
computation graph consisting of 12 layers and 12 heads, we were able to identify prominent patterns in a subset of the layers.

\begin{figure*}[t]
    \begin{subfigure}{.33\textwidth}
        \centering
        \includegraphics[width=.9\linewidth]{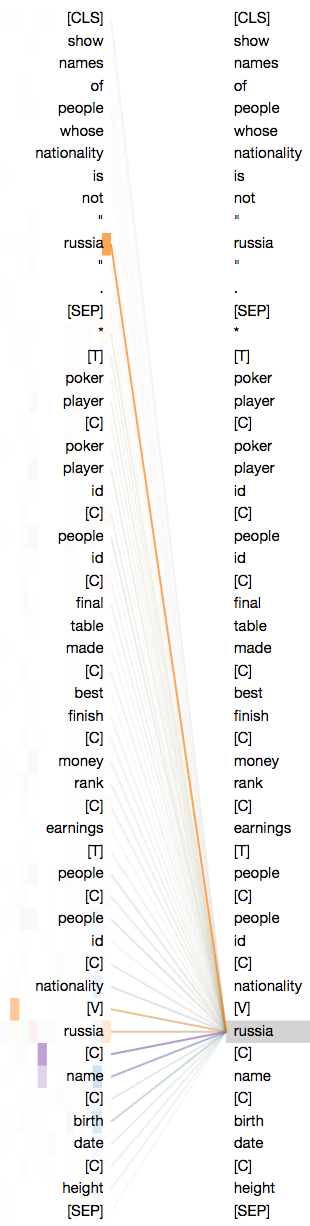}
        \caption{Layer = 2}
    \end{subfigure}
    \begin{subfigure}{.33\textwidth}
        \centering
        \includegraphics[width=.9\linewidth]{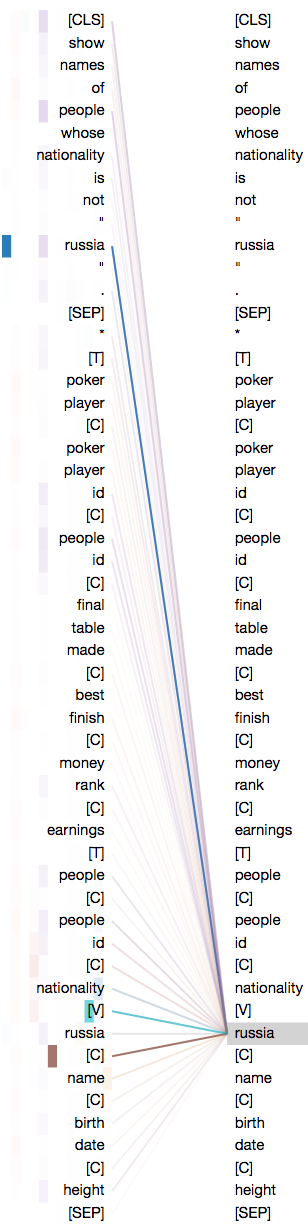}
        \caption{Layer = 4}
    \end{subfigure}
    \begin{subfigure}{.33\textwidth}
        \centering
        \includegraphics[width=.9\linewidth]{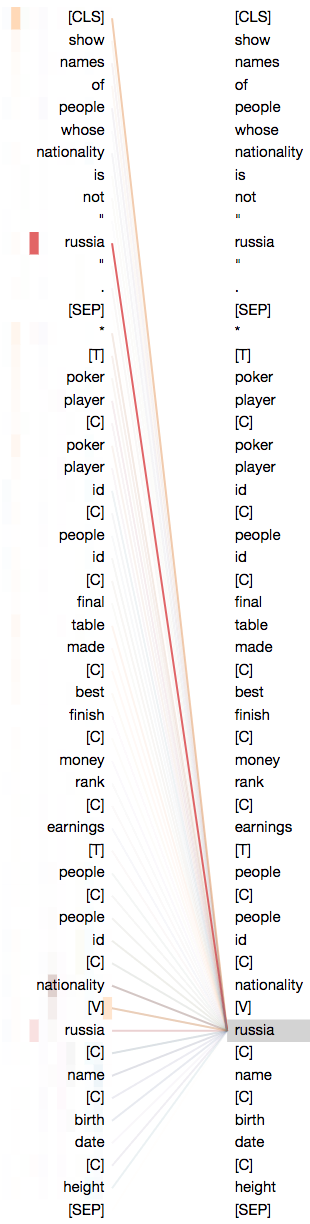}
        \caption{Layer = 5}
    \end{subfigure}
    \caption{Visualization of attention to anchor text ``Russia'' from other words. In the shown layers, weights from the textual mention ``Russia'' is significantly higher than the other tokens.}
    \label{fig:russia_vis}
\end{figure*}

\begin{figure*}[t]
    \begin{subfigure}{.5\textwidth}
        \centering
        \includegraphics[width=.6\linewidth]{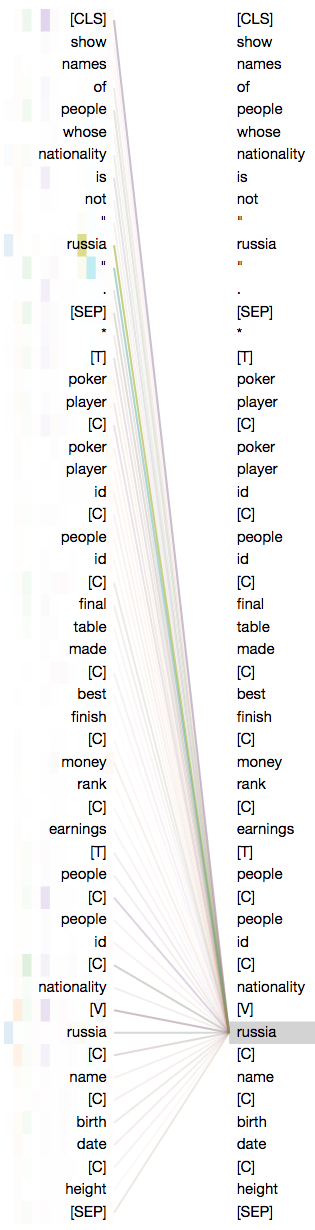}
        \caption{Layer = 10}
    \end{subfigure}
    \begin{subfigure}{.5\textwidth}
        \centering
        \includegraphics[width=.6\linewidth]{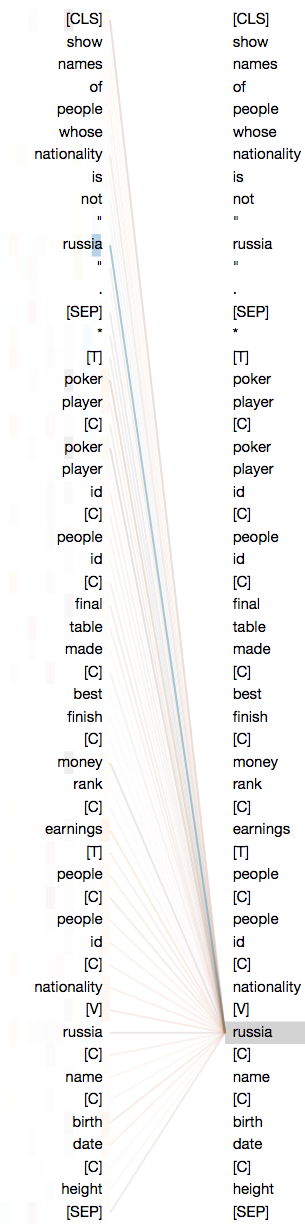}
        \caption{Layer = 11}
    \end{subfigure}
    \caption{Visualization of attention to anchor text ``Russia'' from other words. Continue from Figure~\ref{fig:russia_vis}.}
    \label{fig:russia_vis_2}
\end{figure*}

First, we examine if anchor texts indeed have the effect of bridging information across the textual and tabular segments. The example question we use is ``show names of people whose nationality is not Russia'' and ``Russia'' in the field \<People.Nationality> is identified as the anchor text. As show in Figure~\ref{fig:russia_vis} and Figure~\ref{fig:russia_vis_2}, we find strong connection between the anchor text and their corresponding question mention in layer 2, 4, 5, 10 and 11.

\begin{figure*}[t]
    \begin{subfigure}{.5\textwidth}
        \centering
        \includegraphics[width=.6\linewidth]{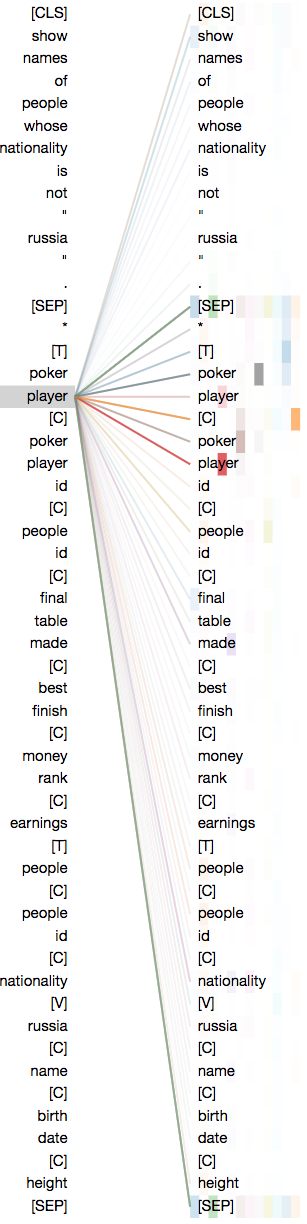}
        \caption{Table \<Poker\_Player>}
    \end{subfigure}
    \begin{subfigure}{.5\textwidth}
        \centering
        \includegraphics[width=.6\linewidth]{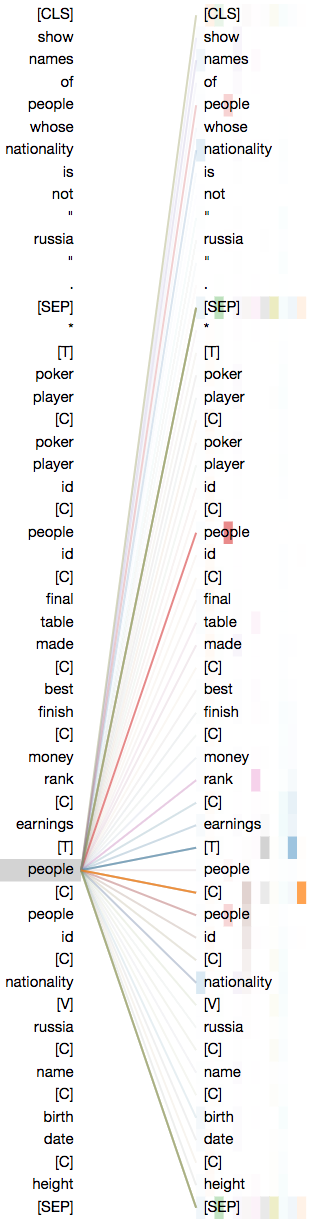}
        \caption{Table \<People>}
        \label{fig:people_pri_vis}
    \end{subfigure}
    \caption{Visualization of attention in layer 5 from tables to their primary keys. In Figure~\ref{fig:people_pri_vis}, the table name \<People> has high attention weights to \<Poker\_Player.People\_ID>, a foreign key referring to its primary key \<People.People\_ID>.}
    \label{fig:table_structure_vis_pri}
\end{figure*}

\begin{figure*}[t]
    \begin{subfigure}{.5\textwidth}
        \centering
        \includegraphics[width=.6\linewidth]{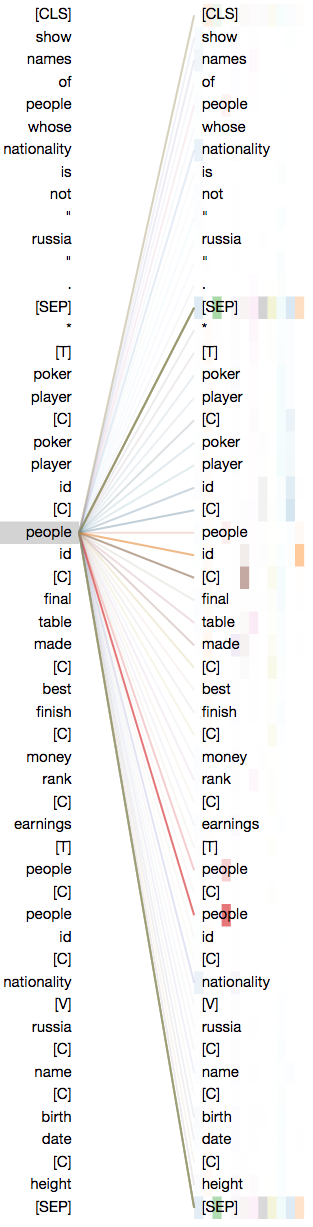}
        \caption{\<Poker\_Player.People\_ID> $\rightarrow$ \<People.People\_ID>}
    \end{subfigure}
    \begin{subfigure}{.5\textwidth}
        \centering
        \includegraphics[width=.6\linewidth]{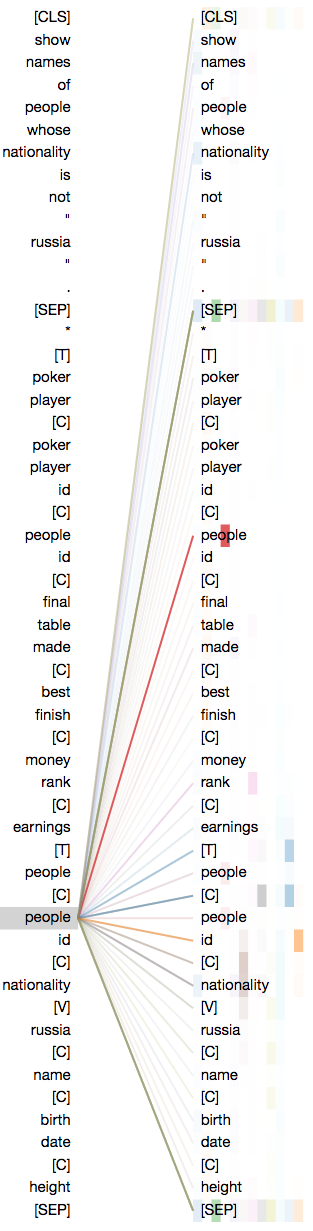}
        \caption{\<People.People\_ID> $\rightarrow$ \<Poker\_Player.People\_ID>}
    \end{subfigure}
    \caption{Visualization of attention in layer 5 between a pair of foreign keys.}
    \label{fig:table_structure_vis_for}
\end{figure*}

We further notice that the layers effectively captures the relational DB structure. As shown in Figure~\ref{fig:table_structure_vis_pri} and Figure~\ref{fig:table_structure_vis_for}, we found attention patterns in layer 5 that connect tables with their primary keys and foreign key pairs.

\begin{figure*}[t]
    \begin{subfigure}{.5\textwidth}
        \centering
        \includegraphics[width=.6\linewidth]{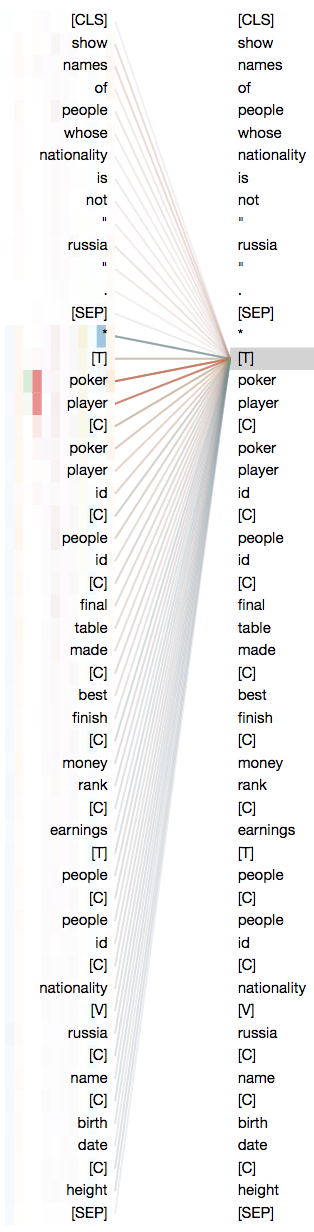}
        \caption{}
    \end{subfigure}
    \begin{subfigure}{.5\textwidth}
        \centering
        \includegraphics[width=.6\linewidth]{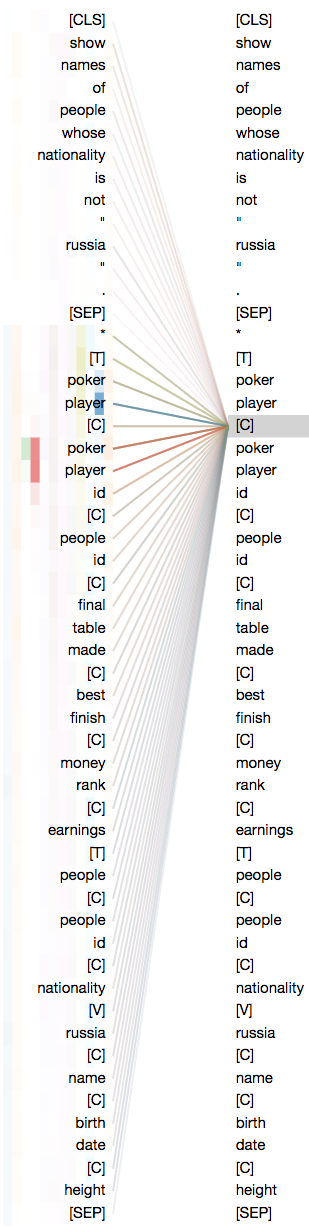}
        \caption{}
    \end{subfigure}
    \caption{Visualization of attention over special tokens \<[T]> and \<[C]> in layer 1.}
    \label{fig:pooling}
\end{figure*}

We notice that all interpretable attention connections are between lexical items in the input sequence, not including the special tokens (\<[T]>, \<[C]>, \<[V]>). This is somewhat counter-intuitive as the subsequent layers of \Ours use the special tokens to represent each schema component. We hence examined attention over the special tokens (Figure~\ref{fig:pooling}) and found that they function as bindings of tokens in the table names and field names. The pattern is especially visible in layer 1. As shown in Figure~\ref{fig:pooling}, each token in the table name ``poker player'' has high attention to the corresponding \<[T]>. Similarly, each token in the field name ``poker player ID'' has high attention to the corresponding \<[C]>. We hypothesize that this way the special tokens function similarly as the cell pooling layers proposed in TaBERT~\cite{DBLP:journals/corr/abs-2005-08314}.

\subsection{Future Improvements}
\label{sec:discussion}

We discuss a few aspects of \Ours that can be improved in future work. 

\paragraph{Anchor Selection} \Ours adopts simple string matching for anchor text selection. In our experiments, improving anchor text selection accuracy significantly improves the end-to-end accuracy. Extending anchor text matching to cases beyond simple string match (e.g. ``LA''$\rightarrow$``Los Angeles'') is a future direction. Furthermore, this step can be learned either independently or jointly with the text-to-SQL objective. Currently \Ours ignores number mentions. We may introduce features which indicate a specific number in the question falls within the value range of a specific column. 

\paragraph{Input Size} As \Ours serializes all inputs into a sequence with special tags, a fair concern is that the input would be too long for large relational DBs. We believe this can be addressed with recent architecture advancements in transformers~\cite{DBLP:journals/corr/abs-2004-05150}, which have scaled up the attention mechanism to model very long sequences.

\paragraph{Relation Encoding} \Ours fuses DB schema meta data features to each individual table field representations. This mechanism loses some information from the original graph structure. It works well on Spider, where the foreign key pairs often have exactly the same names. We 
consider regularizing 
a subset of the attention heads~\cite{DBLP:conf/emnlp/StrubellVAWM18} to capture DB connections a promising way to model the graph structure of relational DBs within the \Ours framework without introducing (a lot of) additional parameters.

\end{document}